\documentclass{article}
\usepackage{amssymb}
\usepackage{amsmath}
\usepackage{graphicx}
\usepackage{subcaption}
\usepackage{booktabs}
\usepackage{multirow}
\usepackage{bm}
\usepackage{wrapfig}

\usepackage{xspace}
\newif\ifshowcomments
\showcommentstrue
\ifshowcomments
    \usepackage[textsize=scriptsize,textwidth=3.2cm]{todonotes}
\else
    \usepackage[disable,textsize=scriptsize,textwidth=3.2cm]{todonotes}
\fi

\definecolor{lightblue}{rgb}{0.3,0.7,1}

\newcommand{\alg}{\textsc{CSA}\xspace}
\newcommand{\edm}{\textsc{EDM}\xspace}

\newcommand{\Cond}{\textsc{$\textrm{CSA}$}\xspace}
\newcommand{\JointCond}{\textsc{$\textrm{CSA}^\dagger$}\xspace}
\usepackage{xspace}
\usepackage[preprint]{corl_2025} %

\title{FlashBack: Consistency Model-Accelerated\\ Shared Autonomy}

\author{
  Luzhe Sun, Jingtian Ji, Xiangshan Tan, Matthew R. Walter\\
  Toyota Technological Institute at Chicago, 
  United States\\
  \texttt{\{luzhesun, jijingtian, 
vincenttann, 
mwalter\}@ttic.edu} \\
}

\begin{document}

\maketitle
\vspace{-5mm}
\begin{abstract}
    Shared autonomy is an enabling technology that provides users with control authority over robots that would otherwise be difficult if not impossible to directly control. Yet, standard methods make assumptions that limit their adoption in practice---for example, prior knowledge of the user's goals or the objective (i.e., reward) function that they wish to optimize, knowledge of the user's policy, or query-level access to the user during training. Diffusion-based approaches to shared autonomy do not make such assumptions and instead only require access to demonstrations of desired behaviors, while allowing the user to maintain control authority. However, these advantages have come at the expense of high computational complexity, which has made real-time shared autonomy all but impossible. To overcome this limitation, we propose Consistency Shared Autonomy (\alg), a shared autonomy framework that employs a consistency model-based formulation of diffusion. Key to \alg is that it employs the distilled probability flow of ordinary differential equations (PF
    ODE) to generate high-fidelity samples in a \emph{single} step. This results in inference speeds significantly than what is possible with previous diffusion-based approaches to shared autonomy, enabling real-time assistance in complex domains with only a single function evaluation. Further, by intervening on flawed actions at intermediate states of the PF ODE, \alg enables varying levels of assistance. We evaluate \alg on a variety of challenging simulated and real-world robot control problems, demonstrating significant improvements over state-of-the-art methods both in terms of task performance and computational efficiency. Our code is available at \url{https://ripl.github.io/CSA-website/}.
\end{abstract}


\section{Introduction}
\label{sec:intro}

Shared autonomy is a collaborative control paradigm where a human operator and an autonomous agent jointly control a robotic system to achieve common objectives~\cite{tothenoiseandback,MDP-shared-auto, Shared-autonomy-concept1, Shared-autonomy-concept2, Shared-autonomy-concept3}. By complementing human intuition with machine precision, shared autonomy enhances performance, ensures safety, and reduces operator workload. This is particularly valuable in complex control scenarios where it enables a human \textbf{pilot} to provide high-level input while the robotic \textbf{copilot} autonomously manages low-level motion corrections to maintain safety and operational efficiency~\cite{sha-only-when-necessary, deepRL-sha, residule-sha, ghorbel2018decision, schroer2015autonomous,underwater-sha-auto}.

Traditionally, shared autonomy algorithms have assumed the existence of a fixed set of known goals from which the user's goal is be inferred at test time based on their input~\cite{goal-cond-1,goal-cond-2,goal-cond-3,goal-cond-4}. While effective in some settings, these approaches struggle in unstructured and semi-structured settings where the space of goals is not well defined. Recent advancements in generative modeling, particularly diffusion probabilistic models that have revolutionized domains such as image, audio, and 3D generation~\cite{DDPM, DDIM, EDM, DiffWave, image-gen, diff-audio-gen, diff-image-editing, 3Ddiffusion, xu2024set, sun2024stackgen}, offer a promising avenue to address these challenges by framing shared autonomy as sampling from a learned distribution over actions~\cite{diffusionpolicy, tothenoiseandback,motion-planning-diffusion}.

In shared autonomy specifically, diffusion-based methods have demonstrated effectiveness in stabilizing user actions without explicit assumptions regarding the goal space. For instance, \citet{tothenoiseandback} introduced partial diffusion with noise injection to maintain user intention, and \citet{dexteritygen} utilized motion gradients as denoising guidance for context-rich manipulation tasks. However, deploying diffusion models in real-time robotics applications remains challenging due to the computational cost of the diffusion process.
 
Diffusion-denoising probabilistic models (DDPMs)~\cite{DDPM} solve a reverse-time stochastic differential equation (SDE) through a denoising process, however this process typically requires hundreds of steps. This can lead to high time complexity for generation/inference, which precludes real-time deployment. Further, DDPMs employed for shared autonomy are prone to generating samples that may ignore information provided by the user. Thus, real-time shared autonomy faster generative methods that respect the user's input.

\begin{figure}[!t]
    \centering
    \includegraphics[width=1.0\linewidth]{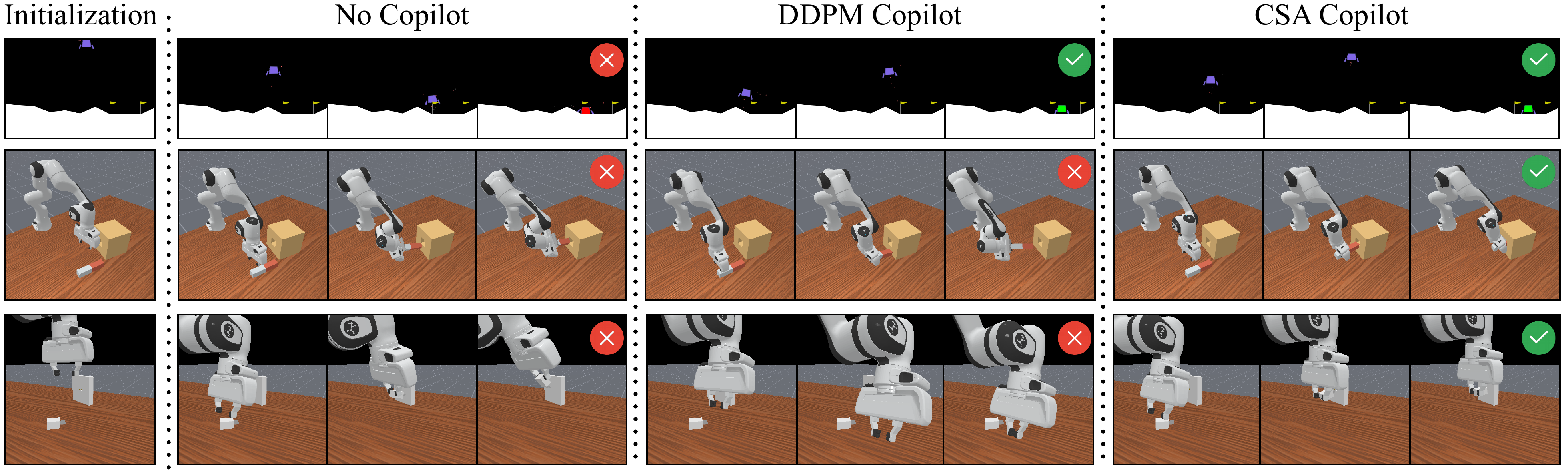}
    \caption{A visualization of the result of using our proposed Consistency Shared Autonomy (CSA) algorithm in comparison to the state-of-the-art DDPM-based shared autonomy baseline on three challenging control tasks.}
    \label{fig:quick-result}
    \vspace{-18pt}
\end{figure}
Consistency models (CM)~\cite{CMmodel} have emerged as a compelling solution to these concerns. Rather than iteratively removing noise as in traditional diffusion models, CM directly learns a denoising function to ``jump'' efficiently to high-quality results. Leveraging a pretrained ordinary differential equation (ODE)-based probability flow from EDM models~\cite{EDM}, CM effectively distills the complex ODE flow into rapid, one-step (or few-step) generative processes without sacrificing fidelity. Empirical successes in image synthesis demonstrate the resulting improvements in efficiency. Notably, \citet{CMmodel} demonstrates state-of-the-art image generation performance using significantly fewer sampling iterations compared to standard diffusion methods. Similarly, consistency policies applied in robotic planning have achieved inference speeds up to 100 times faster than traditional DDPM-based approaches while maintaining competitive performance~\cite{ConsistencyPolicy,diffusionpolicy}.

Motivated by the use of partial diffusion as a means of balancing user control with task performance~\cite{tothenoiseandback,SDEdit} and the computational efficiency of ODE-based distillation methods~\cite{CMmodel}, we propose the Consistency Model for Shared Autonomy (\alg, illustrated in Fig.~\ref{fig:quick-result}). Unlike contemporary shared autonomy methods, \alg provides accelerated inference, ensures proximity to nearest expert actions, and requires minimal training data. We evaluate \alg on a series of simulated and real-world control tasks, demonstrating its improvements over the previous state-of-the-art.

\vspace{-10pt}
\section{Related Work} \label{sec:relatedwork}
\vspace{-10pt}

Early approaches to shared autonomy assumed prior knowledge of the user’s goal~\cite{early-sha1, early-sha2} or that it could be explicitly inferred from user input~\cite{argall2015modular, goal-cond-1, goal-cond-2, goal-cond-3, goal-cond-4}. \citet{deepRL-sha} introduced model-free deep reinforcement learning (RL) for shared autonomy, removing the requirement for known environment dynamics; subsequent works further developed this approach~\cite{residule-sha, tan2022optimizing}. Diffusion model-based action planning has proven powerful in policy learning~\cite{diffusionpolicy, diffuser, ConsistencyPolicy} as well as  shared autonomy. \citet{tothenoiseandback} propose a \emph{partial diffusion} mechanism that adds noise to the user’s action and then employs a partial reverse diffusion process to refine the action towards the training distribution, thereby preserving the user’s intent (\emph{fidelity}) while correcting the action (\emph{conformity}). 
However, in such an SDE-based model, the injection of fresh noise at each reverse step is a double-edged sword: it increases diversity but can push the outcome into undesirable modes. DexterityGen (DexGen)~\cite{dexteritygen} takes a different approach, forgoing partial diffusion in favor of a \emph{user motion promotion} strategy to find a trajectory that aligns with the user’s intention. 

Despite these advances, both of the above diffusion-based approaches (using either a DDPM or DDIM formulation) are constrained by iterative inference. In contrast, \alg uses a one-step inference paradigm that enables microsecond-scale generation, requires only a small amount of training data, does not rely on explicit user goal prediction, yet is performant in high-precision tasks.
 
Compared to the standard formulation of diffusion (e.g., DDPM~\cite{DDPM}) discussed above, Denoising Diffusion Implicit Models (DDiM)~\cite{DDIM} recast the reverse process as a deterministic ordinary differential equation (ODE). EDM~\cite{EDM} unifies SDE and ODE sampling through improved preconditioning and step weighting. 
Notably, ODE-based sampling yields samples to their nearest neighbor along the ideal denoising path, a guarantee that SDE-based diffusion lacks. However, aggressively reducing the number of steps can still degrade output quality despite faster sampling. 

Distillation techniques have been explored to further mitigate inference costs~\cite{meng2023distillation, distill-image-gen}. These approaches typically involve training an ODE-based \emph{teacher model} to generate detailed, high-quality trajectories, followed by training a \emph{student model} that learns to approximate these trajectories with fewer intermediate steps. For instance, Consistency Models (CM), proposed by \citet{CMmodel}, leverage the inherent self-consistency of ODE trajectories. Given two distinct intermediate states $\{x^u,x^v\}$ along the same trajectory, the CM enforces predictions to converge to an identical clean target $\hat{x}_0$. By optimizing this consistency, CMs facilitate rapid sampling in as few as one or several inference steps, significantly improving efficiency without sacrificing generation quality.
\vspace{-5pt}
	
\section{Method}
\label{sec:method}
\vspace{-5pt}

\begin{wrapfigure}{r}{0.45\linewidth}
    \vspace{-2pt}
    \includegraphics[width=\linewidth]{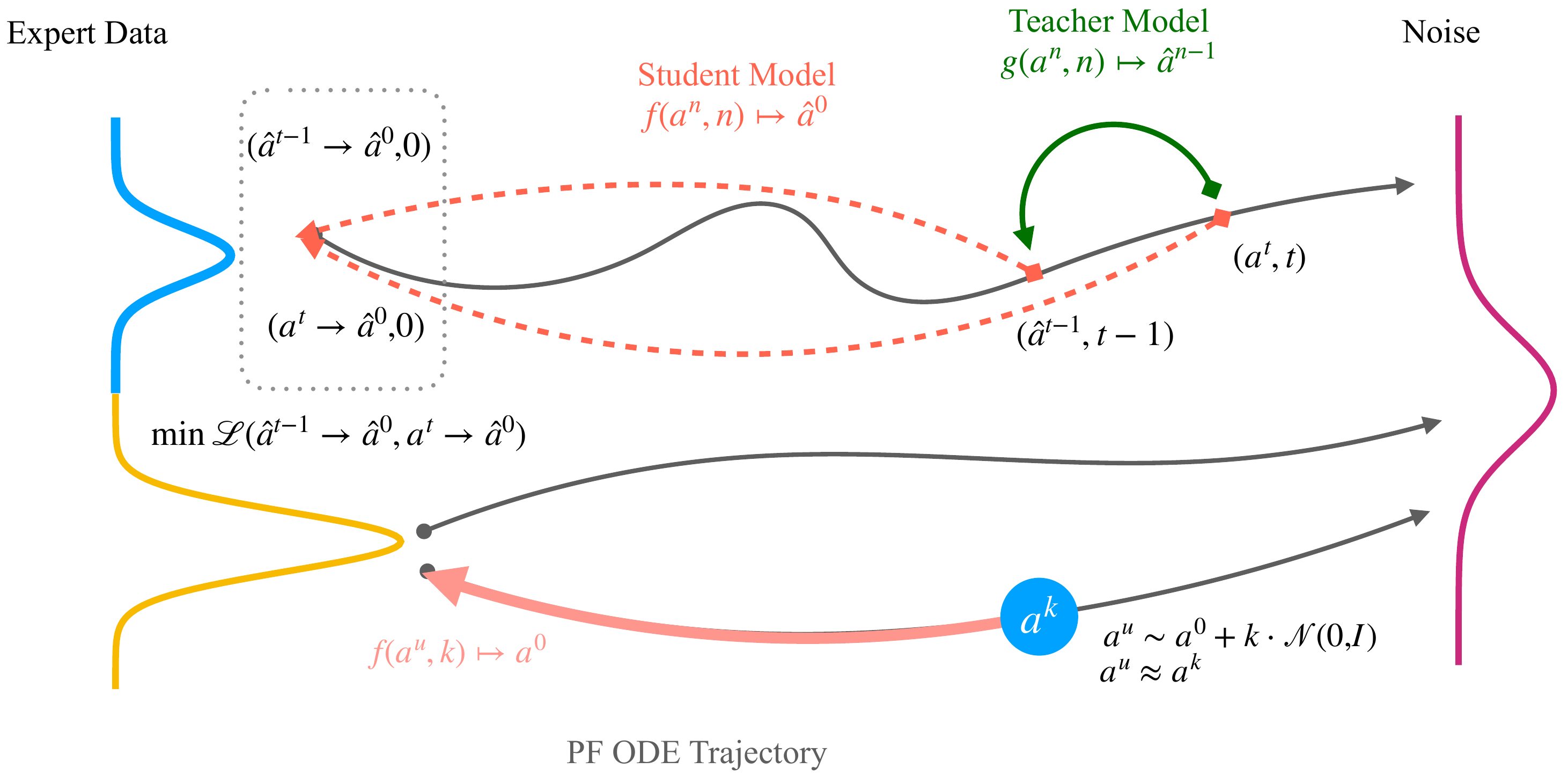}
      \caption{Distillation of PF ODE flow: Select two distinct states $\{a^t,a^{t-1}\}$ along the same trajectory, the CM enforces that predictions converge to the same target $\hat{a}^0$.}
      \label{fig:distill-illustration}
\end{wrapfigure}
Underlying our proposed method is the formulation of shared autonomy as a generative process, whereby we sample actions from a learned expert distribution that are consistent with user's latent intent. Diffusion models provide a compelling way to generate these samples since user-provided actions can be thought of as noisy inputs that lie on the trajectory swept via reverse diffusion (we refer the reader to Appendix~\ref{sec:preliminary-pf-odes} for a discussion of probability flow ODEs). This is the approach that we adopt here. However, as we have previously discussed, the computational cost of this reverse diffusion process precludes real-time operation critical to shared autonomy. 

Rather than relying on many sequential denoising steps, \alg uses consistency-model distillation to collapse the entire ODE-based diffusion trajectory into a single, efficient step. Concretely, we first train a high-fidelity \emph{teacher} diffusion model by solving the PF ODE iteratively, a process that produces reliable samples but requires dozens or more solver steps. During distillation, we then train a \emph{student} model to learn a direct mapping from the teacher’s initial noisy input all the way to its final clean output. In effect, the student ``jumps'' over the intermediate flow trajectory in one shot, approximating the teacher’s multi-step refinement with a single forward pass (or very few steps) while retaining comparable sample quality.

\vspace{-5pt}
\subsection{Training Phase}
\label{sec: training}
\vspace{-5pt}
We train our \alg model by first integrating state prediction into the EDM framework to serve as a teacher model. Subsequently, the consistency model (CM)–based denoiser \alg distills ODE‐flow knowledge from this teacher to perform one‐step denoising, substantially accelerating inference. 

\begin{figure}[!t]
    \centering
    \includegraphics[width=0.95\linewidth]{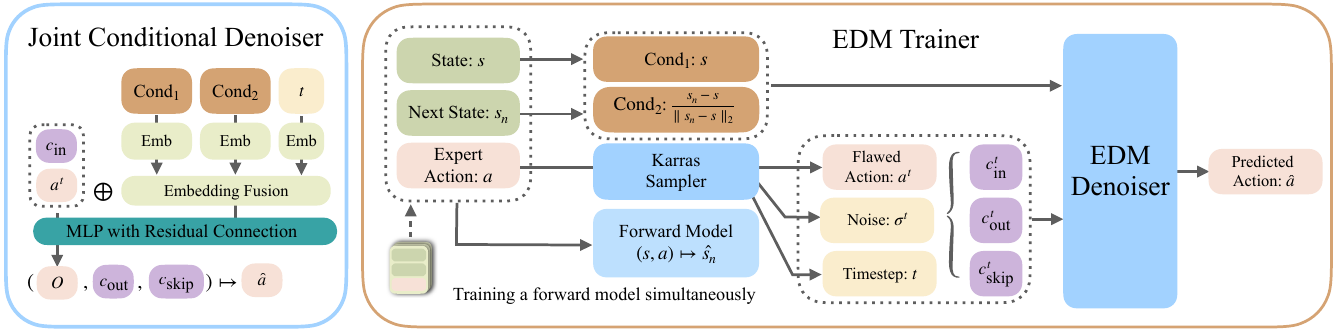}
    \caption{Training Process of EDM (teacher) model.}
    \label{fig:teacher_training}
    \vspace{-10pt}
\end{figure}

\vspace{-5pt}
\paragraph{Teacher Model}
Training data consists of state-action-next state transitions $(s,a,s_n)$, where $s$ is the current state,
$a$ is the expert action, and $s_n$ is the next state. 
\begin{figure}[!t]
    \centering
    \includegraphics[width=0.95\linewidth]{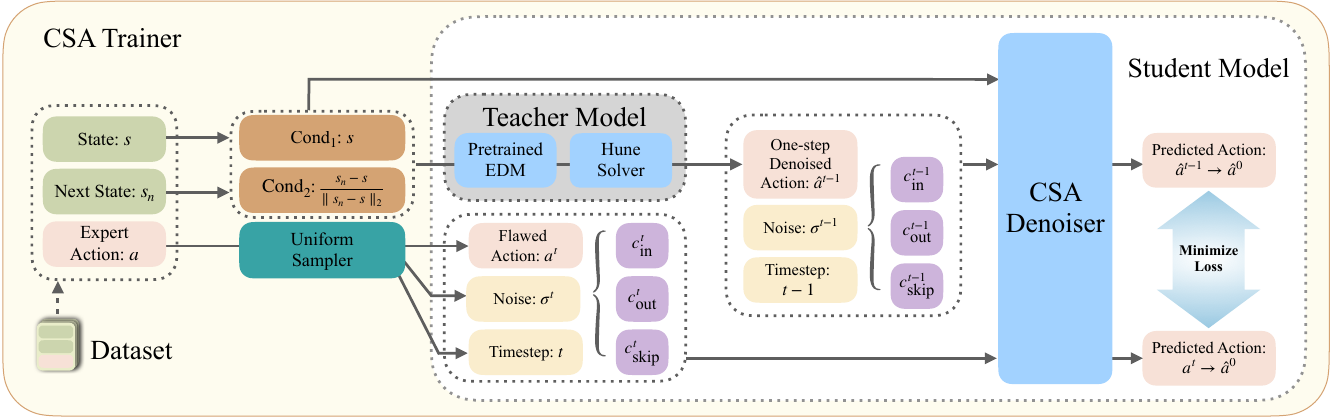}
    \caption{Training Process of CSA (student) model.}
    \label{fig:student_training}
    \vspace{-10pt}
\end{figure}
We preprocess the state into two conditions, $\mathrm{cond}_1=s$ and $\mathrm{cond}_2 = \frac{s_n-s}{\lVert s_n-s \rVert_2}$, where the latter measures the direction in which the state is changing as an indication of near-term user intent. During training, we may condition solely on $\{\mathrm{cond_1}\}$ or jointly on $\{\mathrm{cond_1, cond_2}\}$. Following the EDM framework, we employ the Karras sampler~\cite{EDM} to partition a noise schedule $\{\sigma^0,\dots, \sigma^t\}$.\footnote{Throughout the paper, we use superscripts to denote diffusion steps. In Fig.~\ref{fig:teacher_training}, we label these as noise steps (which are the same as diffusion time steps) to avoid ambiguity.} At noise step $t$, we obtain a perturbed action $a^t$ by injecting noise $\sigma^t\cdot z, z\sim \mathcal{N}(0,1)$. The hyperparameters $\{c_\mathrm{in}, c_\mathrm{out}, c_\mathrm{skip}\}$ are functions of the noise $\sigma^t$ (Appendix~\ref{Appendix: EDM hyperparameter}). We use the joint conditional denoiser $D(\cdot)$ to remove noise (Fig.~\ref{fig:teacher_training}). The model embeds all conditions $\{\mathrm{Cond}_1, \mathrm{Cond}_2,t\}$ using the same hidden dimension as the action embedding, sums them, and then passes them into a three-layer MLP to predict the output $O$. Each layer incorporates conditioning embeddings via residual connections. 

\begin{subequations}
    \begin{align}
        O &= F_{\theta}(c_\mathrm{in}\cdot a^t, t, \mathrm{cond_1}, \textrm{Optional}(\mathrm{cond_2})) \\
        D& = c_\mathrm{out}\cdot O+ c_\mathrm{skip}\cdot a^t 
    \end{align}
\end{subequations}

A learnable adaptive weighting model $\mathrm{\lambda}(\cdot)$ is introduced to track the loss in EDM $\mathcal{L}_{\edm}(\theta) = \mathbb{E}_{(t, a, a^t\mid a)}[\mathrm{dist}(\lambda(\sigma^t),a,D(\cdot))]$, where
\begin{equation}
    \mathcal{L} = \mathrm{dist}(\lambda(\sigma^t),a,D(\cdot)) = e^{\lambda(\sigma^t)}\cdot \lVert (a- D(c_\mathrm{in}, c_\mathrm{out}, c_\mathrm{skip}, a^t, t, \mathrm{cond_1}, \mathrm{cond_2}))\rVert_2 - \lambda(\sigma^t)
\end{equation}
We use conditional loss and joint conditional loss to track the training metric of $\{\mathrm{cond_1}\}$ and $\{\mathrm{cond_1, cond_2}\}$ situation, during loss compute we drop out the $\mathrm{Cond}_2$ with a probability $\gamma$.
\begin{equation}
    \mathcal{L}_{\edm}(\theta) = (1-\gamma)L_{\mathrm{joint-cond}}+\gamma L_{\mathrm{cond}}
\end{equation}

EDM model inference also employs a discretized numerical solver; as a result, its recursive process remains slow.
\vspace{-4pt}
\paragraph{Forward model} At inference, the true next state $s_n$ cannot be observed without executing the predicted action $\hat{a}$. To overcome this, we concurrently train a forward model $\Phi(s,a)\mapsto \hat{s}_n$, implemented as an MLP with normalization layers. 

\vspace{-4pt}
\paragraph{Student Model}

As visualized in Fig.~\ref{fig:distill-illustration}, the distilling training strategy involves finding two samples $\{a^i, a^j\}, i,j\in \{0,\dots,T\}$ on the same PF ODE trajectory, learning a student denoiser $f(a^t,t, o^t)\mapsto \hat{a}^0$, where $o^t$ denotes all other conditions including $\{c_\mathrm{in}^t,c_\mathrm{out}^t,c_\mathrm{skip}^t, \mathrm{cond}_1, \mathrm{cond}_2, \sigma^t\}$. We denote the process of denoising $\{a^i, a^j\}$ as $f(a^i,i, o^i)\mapsto \hat{a}^{i0}$, $f(a^j,j, o^j)\mapsto \hat{a}^{j0}$. Since the trajectory is a deterministic ODE flow, they should trace back to the same starting point. Then, we can formulate the objective as one of minimizing the distance between them, i.e., $\min \mathrm{MSE}(\hat{a}^{i0},\hat{a}^{j0})$.

In practice, we choose $i=j-1$ and use $t$ and $t-1$
in the following expression. The challenge lies in sampling $a^t$ and $a^{t-1}$ from the same flow. 
We start from expert action $a^0$, inject noise $\sigma^t\cdot z$ to get $a^t$. To get $a^{t-1}$ on the same flow we will use the teacher model $g(\cdot)$ (pretrained EDM denoiser) with a Huen solver: $g(a^t, t, o^t) \mapsto \hat{a}^{t-1}$.
The loss function can now be defined as:
\begin{equation}
    \mathcal{L}_{\mathrm{CSA}}(\theta) = \mathrm{MSE} \left(f(a^t, t, o^t), f(\hat{a}^{t-1},t-1, o^{t-1})\right)
\end{equation}

\begin{figure}[t]
  \centering
  \begin{subfigure}[t]{0.44\textwidth}
      \centering
      \includegraphics[width=0.48\linewidth]{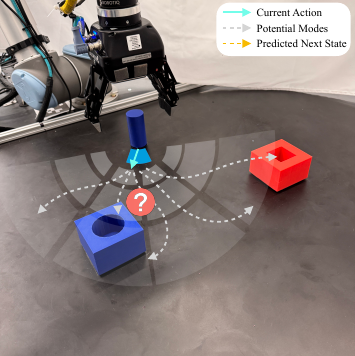}
      \hfill
      \includegraphics[width=0.48\linewidth]{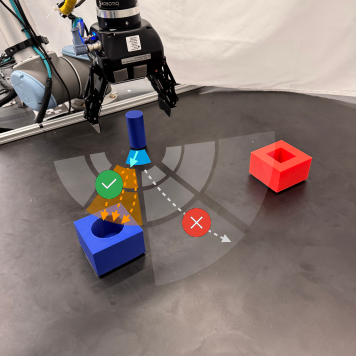}
    \caption{Forward model for next state prediction}
    \label{fig:cm-cmplus}
  \end{subfigure}%
  \hfill
  \begin{subfigure}[t]{0.54\textwidth}
    \centering
    \includegraphics[width=\linewidth]{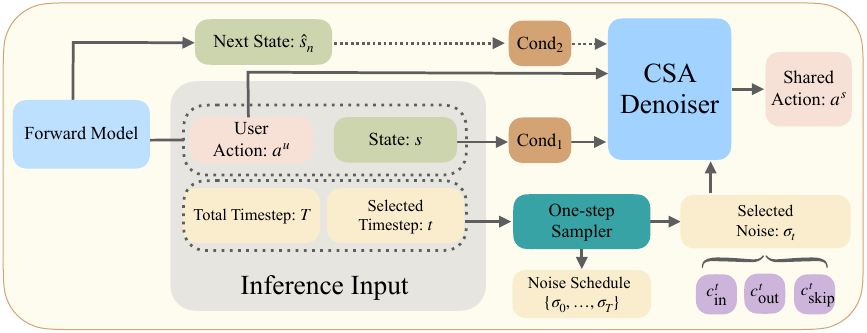}
    \caption{Inference phase of CM model}
    \label{fig:inference}
  \end{subfigure}
\end{figure}
\vspace{-5pt}
\subsection{Inference}\label{sec: Method-Inference} 
\vspace{-5pt}
Unlike previous methods~\cite{tothenoiseandback,ConsistencyPolicy, diffusionpolicy}, the inference phase of \alg does not start from Gaussian noise nor inject noise to the user sample. Given the user's action $a^u$, we assume $a^u \sim \mathcal{N}(a, \sigma^t)$ where $t$ is a hyperparameter determines at which step of the ODE flow the user's action corresponds to (Fig.~\ref{fig:inference}). We define $\alpha = \frac{t}{T}$ which is controlling the amount of noise we assume, this regulates the balance between the fidelity
and the conformity of the generated actions. Using the pretrained forward model, we can get a next state estimation based on user action $a^u$ and current state $s$, $\Phi(a^u, s)\mapsto\hat{s}_n$, providing a short term user intention. Process $\{s,\hat{s}_n\}$ to $\{\mathrm{cond_1}, \mathrm{cond_2}\}$ serve as condition for \alg model. Then we will use One-step Sampler to partition the total noise into $T$ segments, choose noise step $t$ and the noise level $\sigma^t$ that matches $a^u$. Finally apply CM noiser $f(a^u, t, o^t)\mapsto \hat{a}^0$ to finish the one step denosing. We refer to the model conditioned solely on $\{\mathrm{cond_1}\}$ as \Cond, and the variant conditioned on $\{\mathrm{cond_1}, \mathrm{cond_2}\}$ as \JointCond. We show that \JointCond preserves the user’s intent and thereby broadens the effective inference “sweet spot.”

\vspace{-5pt}
\section{Experiments}\label{sec:experiment}
\vspace{-5pt}

We evaluate the performance of our proposed \alg model in simulation on a several continuous control tasks as well as through real-robot experiments.

\subsection{Evaluation Tasks}

We consider the following four continuous control domains as a means of comparing \alg to a contemporary DDPM-based baseline.
Appendix~\ref{sec:appendix-experimental-details} provides further experimental details.

\begin{figure}[!t]
  \centering
  \begin{subfigure}[b]{0.24\textwidth}
    \centering
    \includegraphics[width=\linewidth]{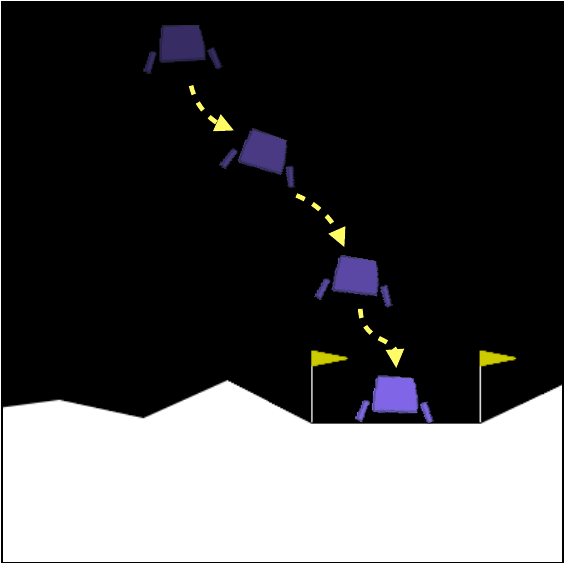}
    \caption{Lunar Lander}
    \label{fig: Lunar Lander}
  \end{subfigure}%
  \hfill
  \begin{subfigure}[b]{0.24\textwidth}
    \centering
    \includegraphics[width=\linewidth]{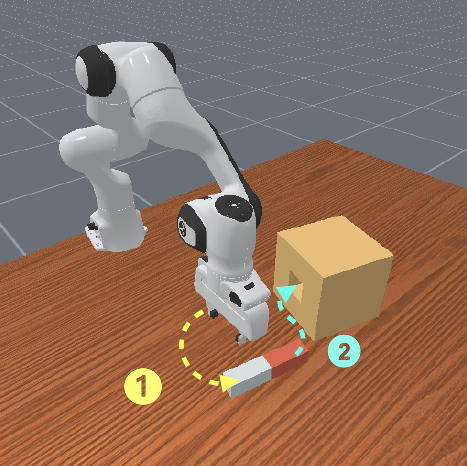}
    \caption{Peg Insertion}
    \label{fig: Peg Insertion}
  \end{subfigure}%
  \hfill
  \begin{subfigure}[b]{0.24\textwidth}
    \centering
    \includegraphics[width=\linewidth]{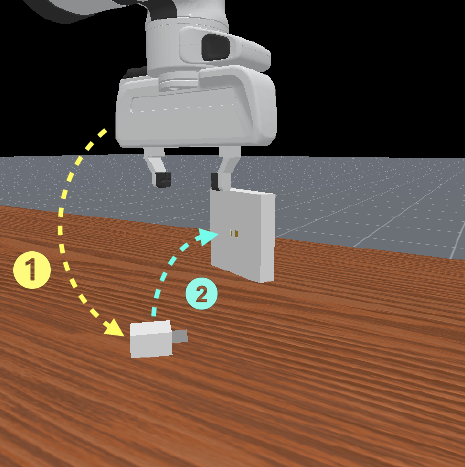}
    \caption{Charger Plug Insertion}
    \label{fig: Charger Plug Insertion}
  \end{subfigure}%
  \hfill
  \begin{subfigure}[b]{0.24\textwidth}
    \centering
    \includegraphics[width=\linewidth]{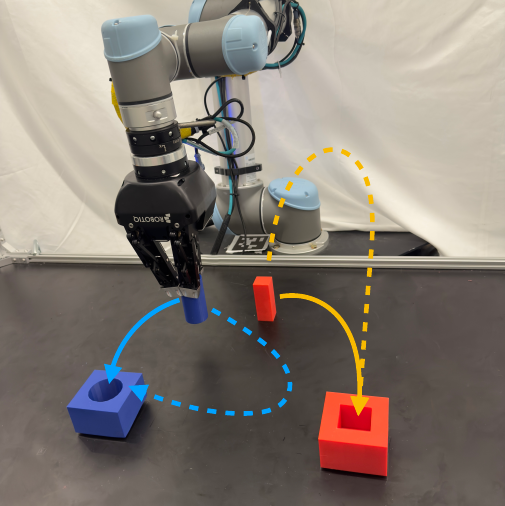}
    \caption{Real Peg Insertion}
    \label{fig: Real Peg Insertion}
  \end{subfigure}
  \caption{Environment Setting}
  \label{fig: Environment Setting}
  \vspace{-15pt}
\end{figure}

\emph{(a) Lunar Lander}: The Lunar Lander environment (Fig.~\ref{fig: Lunar Lander}) is a two-dimensional, continuous-control task adopted from Open AI Gym~\cite{brockman2016openai} that requires landing
a spacecraft on a fixed landing pad. An episode ends when it lands safely, crashes, drifts off course, or when the episode times out after 1000 steps. 

\vspace{-4pt}
\emph{(b) Peg Insertion}: Peg Insertion (Fig.~\ref{fig: Peg Insertion}) tasks a robotic arm with inserting a 10\,cm long, 2\,cm radius peg into a hole that affords only 1\,cm of clearance. The poses of the peg and the target box are randomized at the start of each episode. Success is declared when the peg tip enters the hole more than 1.5\,cm; otherwise the episode terminates after 200 steps. 

\vspace{-4pt}
\emph{(c) Charger Plug Insertion}: Charger Plug Insertion (Fig.~\ref{fig: Charger Plug Insertion}) tightens the tolerance further: a dual-peg charger must slide into its receptacle with only 0.5\,mm of clearance. The poses of the charger and socket are randomized for each episode, and the agent again controls the end-effector using compact kinematic and visual inputs. An episode is determined to be successful if the plug is inserted within 5\,mm and 0.2\,radians of its target pose. Episodes are terminated upon success, when there is a high-impact collision, or if they time out after 300 steps.

\vspace{-4pt}
\emph{(d) Real Peg Insertion}: Real Peg Insertion (Fig.~\ref{fig: Real Peg Insertion}) is performed with a UR5 robot arm equipped with a Robotiq 3-Finger gripper. The task involves inserting either a red square peg or a blue cylinder peg into its corresponding hole, with clearance tolerances of 5\,mm and 4.5\,mm, respectively.

\subsection{Data Collection}
Training the assistive policy requires an oracle operator to provide expert transitions $(s_t, a_t, s_{t+1})$. We use Soft Actor-Critic (SAC)~\cite{SAC} algorithm to train expert policy for Lunar Lander and Peg Insertion, and Proximal Policy Optimization (PPO)~\cite{schulman2017proximal} combined with curriculum learning (CL) to train the expert for Charger Plug Insertion, relaxing the success check to {1.8\,cm then gradually shrinking down to 5\,mm to get an expert for this high precision task~\cite{tang2023industreal}. After we get the expert policy, we apply rejection sampling to collect transits only from successful episodes. All three simulation experiments, regardless of hardness, are trained with $200$K transitions. For the Real Peg Insertion task, demonstrations are collected by experienced human teleoperators using a Meta~Quest~3 VR controller. We collected 180 demonstrations for a total of $62$K transitions in the same form of $(s_t, a_t, s_{t+1})$. This process takes 1 hour with 2 operators.

\subsection{Surrogate Pilots}

While \alg does not require access to a pilot (real or surrogate) for training, we use surrogate pilots in order to scale the number of evaluations. Building off previous work~\cite{deepRL-sha,residule-sha,tothenoiseandback}, we define four different surrogates \textit{noisy}, \textit{laggy}, \textit{noised}, and \textit{slow} as corrupted versions of an expert policy $\pi$.
\begin{equation*}
    \begin{array}{l}
        \operatorname{Noisy}\left(s_t, g\right)=\left\{%
        \begin{array}{ll}
            a_t^e, & \text { if } p\geq \epsilon \\
            a_t^r, & \text { if } p<\epsilon
        \end{array} \quad \operatorname{Laggy}\left(s_t, g\right)=\left\{\begin{array}{ll}
            a_t^e, & \text { if } p\geq \epsilon \\
            a_{t-1}^e, & \text { if } p<\epsilon
        \end{array}\right.\right.\\ \\
        \operatorname{Noised}\left(s_t, g\right)=a_t^e+\mathbb{N}(0, \epsilon) \quad  \quad \quad \operatorname{Slow}\left(s_t, g\right)=(1-\epsilon) \cdot a_t^e
    \end{array}
\end{equation*}
where $\epsilon\in[0,1]$ is a \emph{flaw} parameter that determines the amount of corruption, the subscript $t$ is the timestep (in episode), $a^e_t$ is an expert action, $a^r_t$ is a random action, $s_t$ is the state and $g$ refers to goal. We evaluate four surrogates under different flaw levels $\epsilon$ with 10 random seeds and 30 rollouts each. We provide the parameter $\epsilon$ values in the Appendix \ref{Appendix: surrogate performance} Table \ref{tab:surrogate-flaws}. 

\subsection{Evaluation Results}
\textbf{Simulation Result:}\quad Given that our method is a goal-excluded, state-conditional shared control model, we compare against the SDE-based DDPM baseline of \cite{tothenoiseandback}, adopting their forward diffusion ratio as $\alpha$ to trade off fidelity and conformity. Figures~\ref{fig: Lunar Lander Simulation Result} and~\ref{fig: Peg Insertion Simulation Result} show that in the low-dimensional Lunar Lander task, \alg achieves performance on par with the baseline, while in the higher-dimensional Peg Insertion task, \alg substantially outperforms it—where the DDPM policy yields under a 10\% gain under noise.  

In the high-precision Charger Plug Insertion task, we tested the DDPM baseline across ten noise schedules ($\beta_{\max}\!\sim\!\mathrm{Unif}[0.01,0.25]$) and observed zero successful runs, underscoring its acute sensitivity to schedule tuning and the associated computational cost. By contrast, \alg attains robust success across all tasks with a single hyperparameter setting, eliminating per–task noise tuning and saving considerable time.  

As detailed in Section~\ref{sec: Method-Inference}, the \JointCond variant further preserves user intent at large $\alpha$ by conditioning on the predicted next state. In Fig.~\ref{fig: CSA noised lander},~\ref{fig: CSA+ noised lander} and Fig.~\ref{fig: CSA noisy peg},~\ref{fig: CSA+ noisy peg}, its success rate remains flat beyond the optimal noise ratio, obviating the need for meticulous noise-ratio selection during inference.

\begin{figure}[!t]
  \centering
  \begin{subfigure}[b]{0.33\textwidth}
    \centering
    \includegraphics[width=\linewidth]{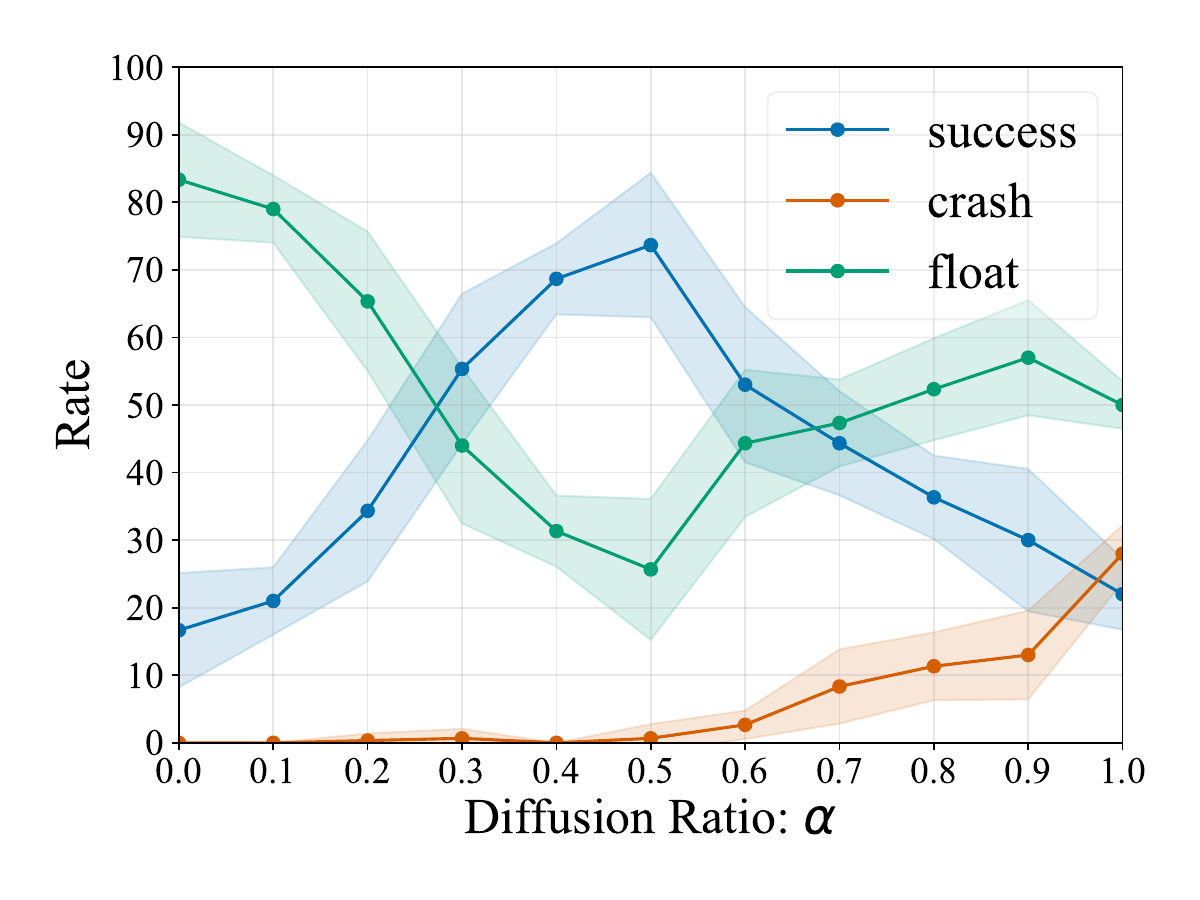}
    \caption{DDPM Policy}
    \label{fig: DDPM noised lander}
  \end{subfigure}%
  \hfill
  \begin{subfigure}[b]{0.33\textwidth}
    \centering
    \includegraphics[width=\linewidth]{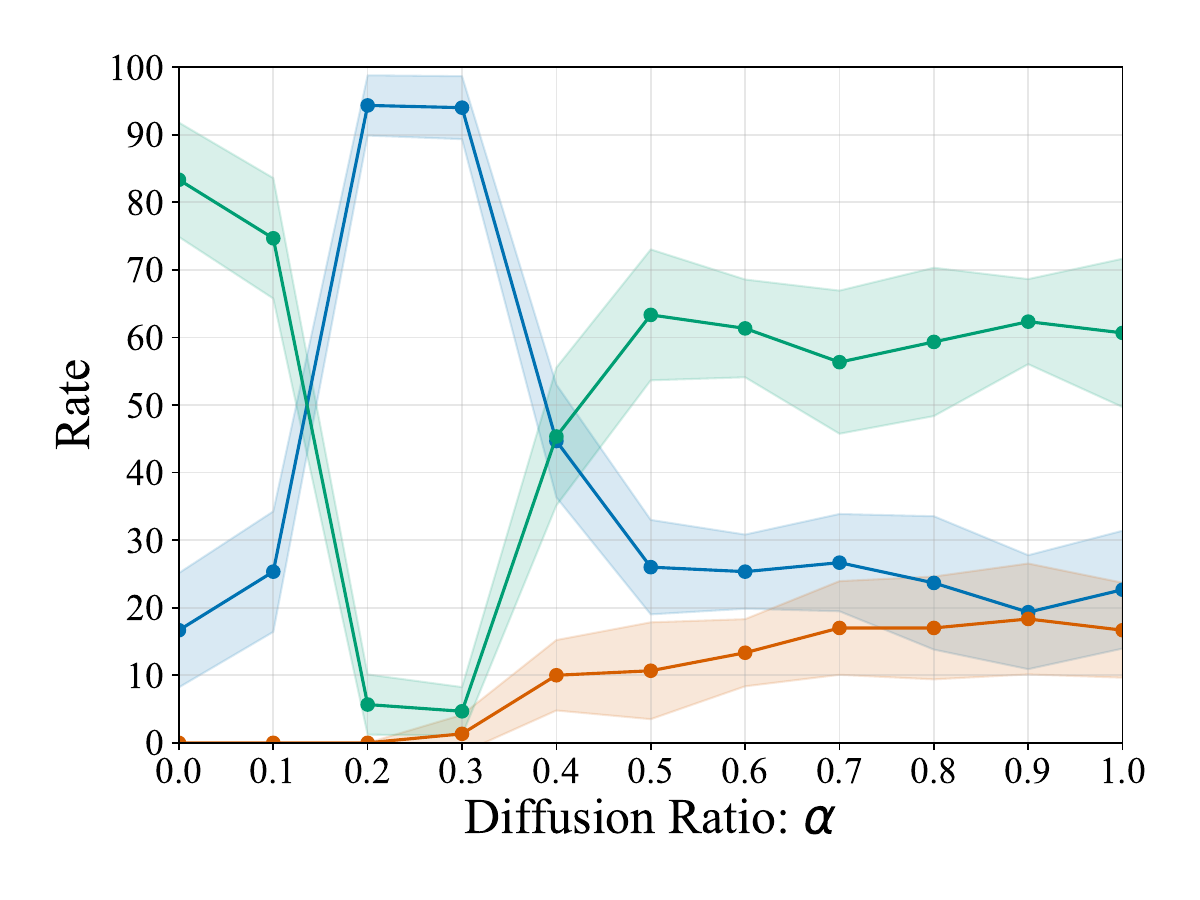}
    \caption{\Cond}
    \label{fig: CSA noised lander}
  \end{subfigure}%
  \hfill
  \begin{subfigure}[b]{0.33\textwidth}
    \centering
    \includegraphics[width=\linewidth]{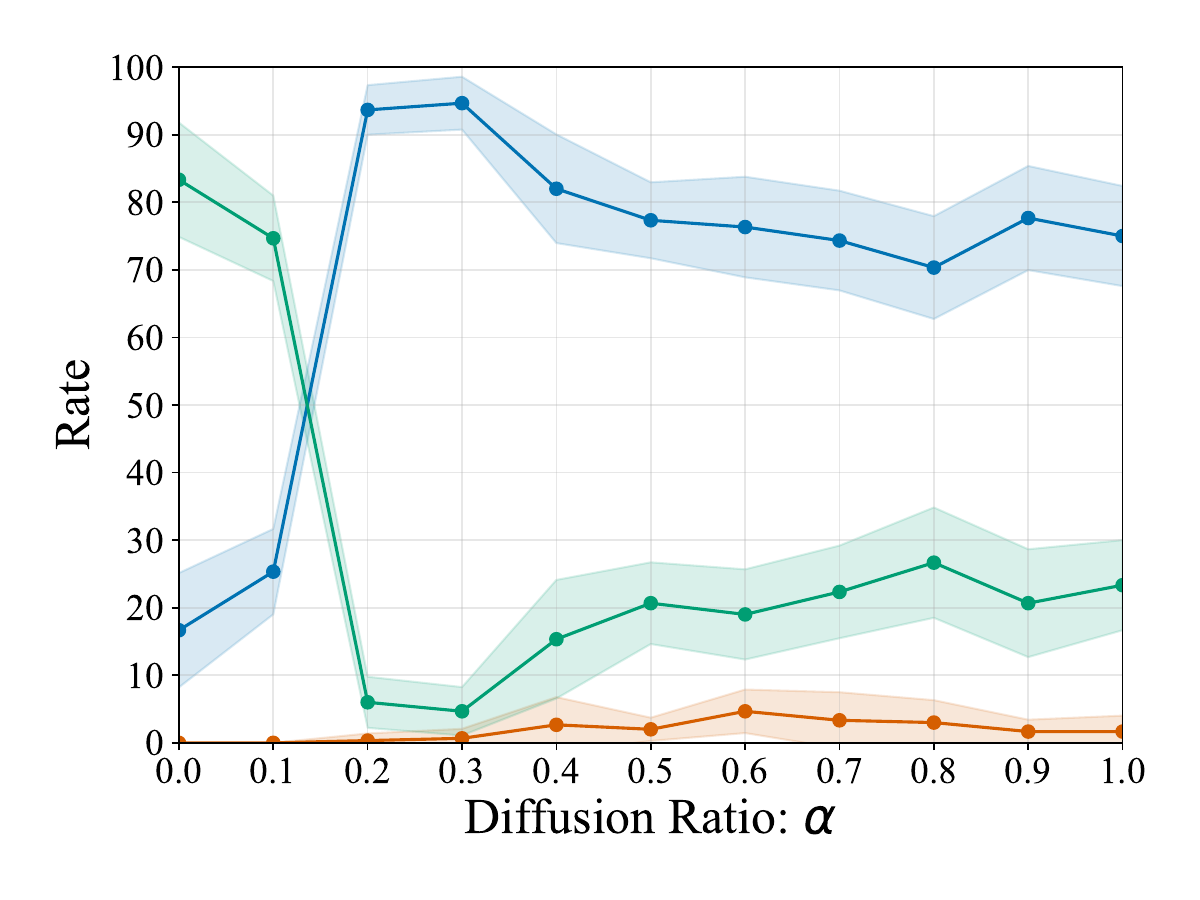}
    \caption{\JointCond}
    \label{fig: CSA+ noised lander}
  \end{subfigure}%
  \caption{Lunar Lander Noised Simulation Result}
  \label{fig: Lunar Lander Simulation Result}
  \vspace{-15pt}
\end{figure}

\begin{figure}[!t]
  \centering
  \begin{subfigure}[b]{0.33\textwidth}
    \centering
    \includegraphics[width=\linewidth]{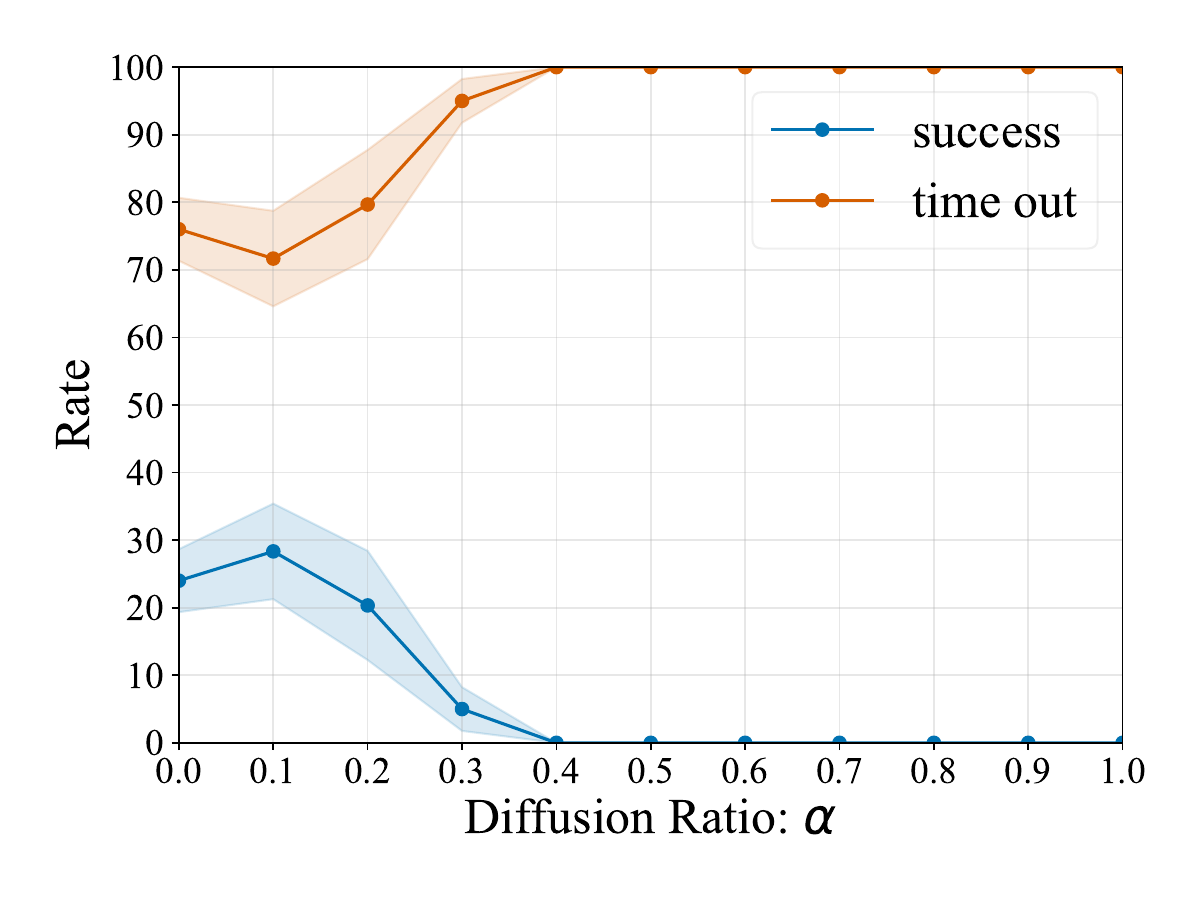}
    \caption{DDPM Policy}
    \label{fig: DDPM noisy peg}
  \end{subfigure}%
  \hfill
  \begin{subfigure}[b]{0.33\textwidth}
    \centering
    \includegraphics[width=\linewidth]{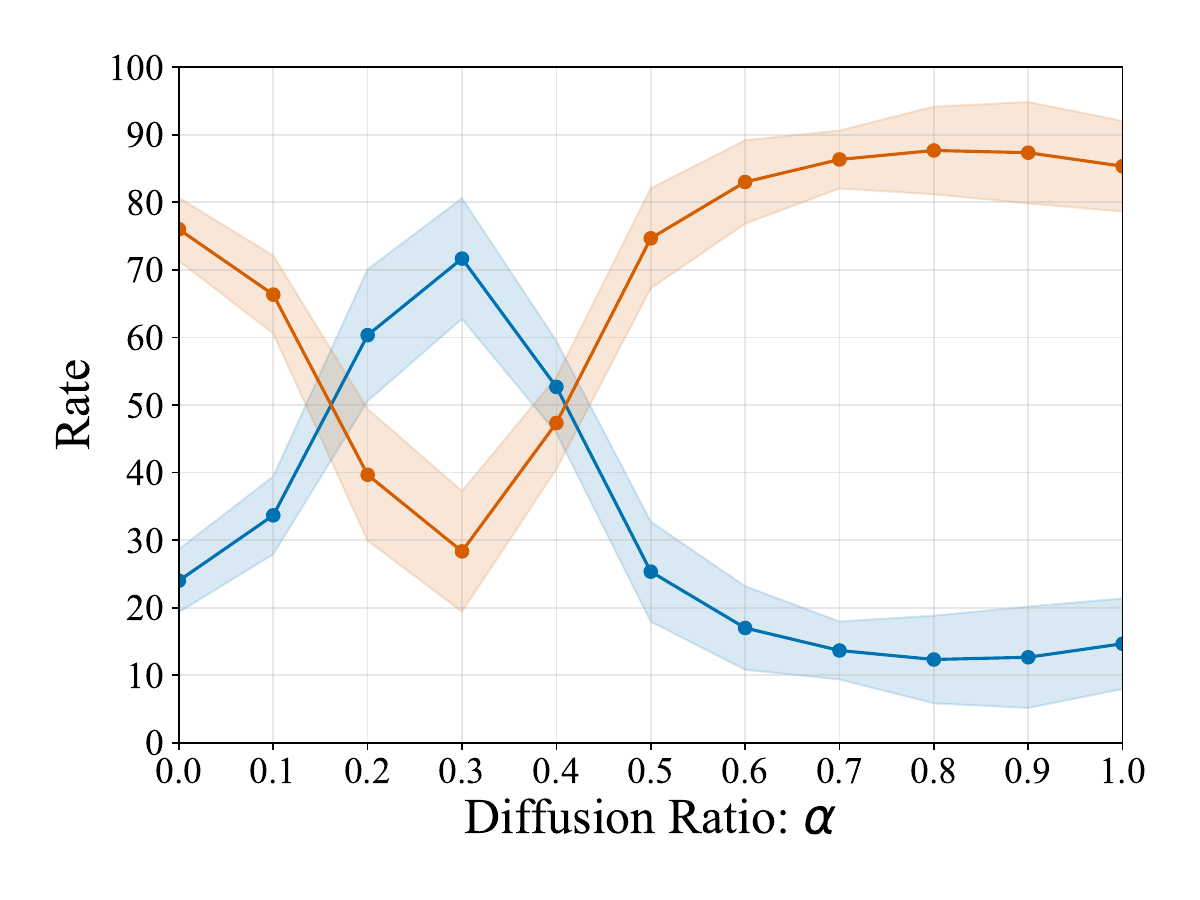}
    \caption{\Cond}
    \label{fig: CSA noisy peg}
  \end{subfigure}%
  \hfill
  \begin{subfigure}[b]{0.33\textwidth}
    \centering
    \includegraphics[width=\linewidth]{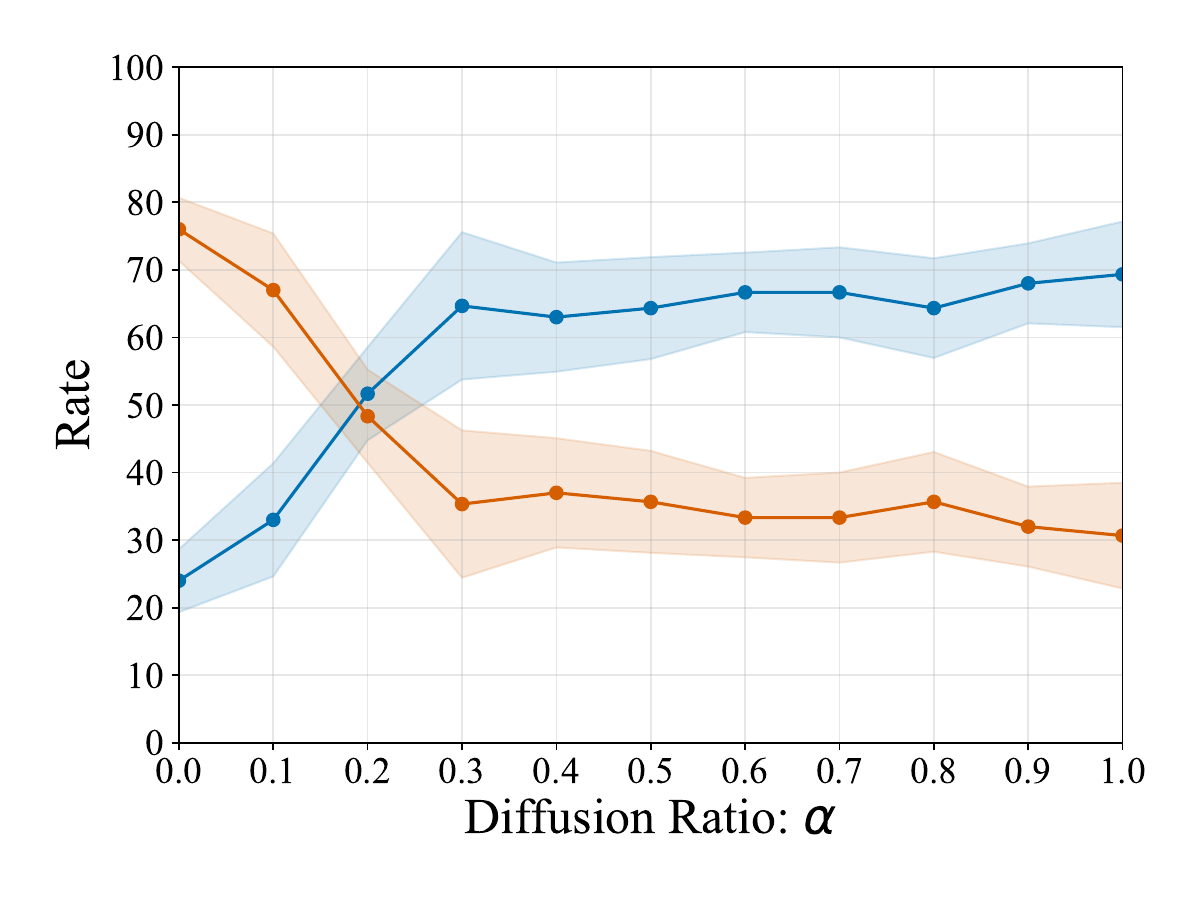}
    \caption{\JointCond}
    \label{fig: CSA+ noisy peg}
  \end{subfigure}%
  \vspace{1.5mm}
  \caption{Peg Insertion Noisy Simulation Result}
  \label{fig: Peg Insertion Simulation Result}
  \vspace{-10pt}
\end{figure}

\begin{table}[!t]
  \centering
  \scriptsize
  \begin{tabular}{%
    l  %
    l  %
    c c c c %
  }
    \toprule
    Actor    & Method    & Success Rate (\%) $\uparrow$ & Crash Rate (\%) $\downarrow$ & NFE & Inference Time (ms) $\downarrow$ \\
    \midrule
    \multirow{4}{*}{Laggy}
      & Surrogate & 40.00 $\pm$ \hphantom{1}4.16 & 52.66 $\pm$ 4.92 & -- & -- \\
      & DDPM      & 75.67  $\pm$ \hphantom{1}9.30 & 14.67 $\pm$ 6.32 & 24 & 13.62 $\pm$ 0.22 \\
      & \Cond        & \textbf{87.67 $\pm$ \hphantom{1}5.89} & \hphantom{1}6.67 $\pm$ 4.16 & 1 & \hphantom{1}0.92 $\pm$ 0.04 \\
      & \JointCond  & \textbf{91.00 $\pm$ \hphantom{1}3.87} & \textbf{\hphantom{1}5.00 $\pm$  3.24} & 1 & \hphantom{1}1.22 $\pm$ 0.30\\
    \midrule
    \multirow{4}{*}{Noisy}
      & Surrogate & 26.00 $\pm$ \hphantom{1}6.44 & \hphantom{1}6.67 $\pm$ 3.85 & -- & -- \\
      & DDPM      & 55.67 $\pm$ \hphantom{1}7.86 & \hphantom{1}4.67 $\pm$ 3.58 & 24 & 17.67 $\pm$ 1.49 \\
      & \Cond        & \textbf{85.00 $\pm$ \hphantom{1}8.05} & \hphantom{1}2.67 $\pm$ 3.44 & 1 & \hphantom{1}1.79 $\pm$ 0.24 \\
      & \JointCond  & \textbf{89.00 $\pm$ \hphantom{1}4.17} & \textbf{\hphantom{1}2.33 $\pm$ 2.25} & 1 & \hphantom{1}2.03 $\pm$ 0.35 \\
    \midrule
    \multirow{4}{*}{Noised}
      & Surrogate & 16.67 $\pm$ \hphantom{1}8.46 & \hphantom{1}0.00 $\pm$ 0.00 & -- & -- \\
      & DDPM      & 73.67 $\pm$ 10.71 & 10.43 $\pm$ 2.11 & 24 & 18.60 $\pm$ 1.51 \\
      & \Cond        & \textbf{94.33 $\pm$ \hphantom{1}4.46} & \textbf{\hphantom{1}0.00 $\pm$ 0.00} & 1 & \hphantom{1}1.94 $\pm$ 0.25 \\
      & \JointCond  & \textbf{94.67 $\pm$ \hphantom{1}3.91} & \hphantom{1}0.67 $\pm$ 1.41 & 1 & \hphantom{1}1.82 $\pm$ 0.59 \\
    \midrule
    \multirow{4}{*}{Slow}
      & Surrogate & 75.33 $\pm$ \hphantom{1}7.57 & 18.33 $\pm$ 5.93 & -- & -- \\
      & DDPM      & \textbf{92.67 $\pm$ \hphantom{1}4.09} & \textbf{\hphantom{1}3.00 $\pm$ 2.92} & 14 & 10.44 $\pm$ 0.07 \\
      & \Cond        & 80.33 $\pm$ \hphantom{1}5.32 & 11.33 $\pm$ 6.13 & 1 & \hphantom{1}1.72 $\pm$ 0.27 \\
      & \JointCond  & 79.00 $\pm$ \hphantom{1}6.10 & 12.67 $\pm$ 6.81 & 1 & \hphantom{1}1.77 $\pm$ 0.24 \\
    \bottomrule
  \end{tabular}
  \vspace{1.5mm}
  \caption{Lunar Lander Performance}
  \label{tab:lunar_lander}
  \vspace{-10pt}
\end{table}

\begin{table}[!t]
  \centering
  \scriptsize
  \begin{tabular}{%
    l  %
    l  %
    c c c  %
    c c c  %
  }
    \toprule
       &        %
      & \multicolumn{3}{c}{Peg Insertion}
      & \multicolumn{3}{c}{Charger Plug Insertion} \\
    \cmidrule(lr){3-5} \cmidrule(lr){6-8}
    Pilot  & Copilot
      & Success Rate(\%) $\uparrow$ & NFE & Inference Time(ms) $\downarrow$
      & Success Rate(\%) $\uparrow$ & NFE & Inference Time(ms) $\downarrow$ \\
    \midrule
    \multirow{4}{*}{Laggy}
      & None & 39.33 $\pm$ 10.16 & -- & -- & 22.00 $\pm$ \hphantom{1}7.57 & -- & -- \\
      & DDPM      & 55.00 $\pm$ \hphantom{1}8.35 & 9 & 5.25 $\pm$ 0.42 & \hphantom{1}0.00 $\pm$ \hphantom{1}0.00 & -- & -- \\
      & \Cond        & \textbf{56.33 $\pm$ 12.22} & 1 & 1.00 $\pm$ 0.19 & \textbf{29.00 $\pm$ 10.89} & 1 & 1.35 $\pm$ 0.37\\
      & \JointCond  & 45.67 $\pm$ \hphantom{1}8.61 & 1 & 1.02 $\pm$ 0.16 & 22.33 $\pm$ \hphantom{1}6.30 & 1 & 1.01 $\pm$ 0.06\\
    \midrule
    \multirow{4}{*}{Noisy}
      & None & 24.00 $\pm$ \hphantom{1}4.66 & -- & -- & 17.00 $\pm$ \hphantom{1}4.57  & -- & -- \\
      & DDPM      & 28.33 $\pm$ \hphantom{1}7.07 & 4 & 2.72 $\pm$ 0.59 & \hphantom{1}0.00 $\pm$ \hphantom{1}0.00 & -- & --\\
      & \Cond        & \textbf{71.67 $\pm$ \hphantom{1}8.92} & 1 & 1.04 $\pm$ 0.15 & 36.67 $\pm$ \hphantom{1}8.31 & 1 & 1.34 $\pm$ 0.41 \\
      & \JointCond  & \textbf{69.33 $\pm$ \hphantom{1}7.83} & 1 & 1.18 $\pm$ 0.31 & \textbf{56.67 $\pm$ \hphantom{1}5.67} & 1 & 1.04 $\pm$ 0.12 \\
    \midrule
    \multirow{4}{*}{Noised}
      & None & 20.00 $\pm$ \hphantom{1}7.03 & -- & -- & 31.67 $\pm$ \hphantom{1}9.06 & -- & -- \\
      & DDPM      & 40.00 $\pm$ \hphantom{1}8.16 & 9 & 5.12 $\pm$ 0.15 & \hphantom{1}0.00 $\pm$ \hphantom{1}0.00 & -- & -- \\
      & \Cond        & \textbf{69.00 $\pm$ \hphantom{1}8.90} & 1 & 1.18 $\pm$ 0.15 & \textbf{70.00 $\pm$ \hphantom{1}8.89} & 1 &  1.59 $\pm$ 0.33 \\
      & \JointCond  & 58.00 $\pm$ \hphantom{1}8.34 & 1 & 1.29 $\pm$ 0.20 & 60.67 $\pm$ \hphantom{1}7.50  & 1 & 1.71 $\pm$ 0.40 \\
    \midrule
    \multirow{4}{*}{Slow}
      & None & 29.67 $\pm$ \hphantom{1}8.38 & -- & -- & 29.33 $\pm$ \hphantom{1}5.40 & -- & -- \\
      & DDPM      & 29.33 $\pm$ \hphantom{1}4.10 & -- & -- & \hphantom{1}0.00 $\pm$ \hphantom{1}0.00 & -- & -- \\
      & \Cond        & \textbf{59.67 $\pm$ 12.32} & 1 & 0.99 $\pm$ 0.18 & 36.33 $\pm$ 10.12 & 1 & 1.25 $\pm$ 0.36 \\
      & \JointCond  & 35.00 $\pm$ \hphantom{1}7.58 & 1 & 1.09 $\pm$ 0.22 & \textbf{50.33 $\pm$ \hphantom{1}8.23} & 1 & 1.25 $\pm$ 0.37 \\
    \bottomrule
  \end{tabular}
  \vspace{1.5mm}
  \caption{Peg Insertion and Charger Plug Insertion}
  \label{tab:peg_charger}
  \vspace{-10pt}
\end{table}

\textbf{Real Robot Result:}\quad We measure the effectiveness of our assistive policy through human user experiments in the Real Peg Insertion environment. We recruited 10 participants (6 identified as male, 3 as female, 1 prefer not to say, with an average age of 25.9).\footnote{None of the participants were co-authors or otherwise involved in this research.} We presented participants with two unidentified copilots, one being our assistive policy \JointCond and the other
being direct teleoperation (i.e., no assistance).
\footnote{DDPM was not included as a baseline in the real-robot experiments because its performance proved extremely sensitive to hyperparameters; despite extensive tuning, we were unable to find any setting that yielded stable convergence on this task.}.\

At the beginning of the experiment, we allowed the participant to practice with both copilots for two trials.
In the subsequent testing phase, every participant controlled the robot arm to insert the square peg and the cylinder peg three times each with one randomly chosen copilot, and then repeated this process with the other copilot.
Participants provided their user actions~$a^u$ through a handheld Quest3 VR controller and viewed the scene only through a side-mounted camera feed, thereby removing normal binocular depth cues. Any trial that exceeded the 60-second limit or triggered an emergency stop due to excessive contact force was marked as a failure.

We consider both the quantitative and qualitative performance of our assistive policy \JointCond. Quantitatively, Table~\ref{tab:real_peg_results} compares the average success rate and the average completion time with and without our assistive policy, showing that providing participants with our assistive policy improved their success rate and reduced the completion time. For qualitative result, we provided it in Appendix \ref{subsec: human-survey-result}.

\begin{table}[!t]
  \centering
  \scriptsize
  \begin{tabular}{l cc cc cc}
    \toprule
    \multirow{2}{*}{Copilot} &
      \multicolumn{2}{c}{Square Peg} &
      \multicolumn{2}{c}{Cylinder Peg} &
      \multicolumn{2}{c}{Overall} \\
    \cmidrule(lr){2-3}\cmidrule(lr){4-5}\cmidrule(lr){6-7}
      & Success~Rate~(\%)~\(\uparrow\) & Time~(s)~\(\downarrow\)
      & Success~Rate~(\%)~\(\uparrow\) & Time~(s)~\(\downarrow\)
      & Success~Rate~(\%)~\(\uparrow\) & Time~(s)~\(\downarrow\) \\ 
    \midrule
None & 60.0 & 25.9 & 73.3 & 30.4 & 66.7 & 28.4 \\
\textsc{\JointCond} (ours) & \textbf{73.3} & \textbf{22.6} & \textbf{93.3} & \textbf{24.3} & \textbf{83.3} & \textbf{24.1} \\
    \bottomrule
  \end{tabular}
  \vspace{1.5mm}
  \caption{Real Peg Insertion performance w/ and w/o our \textsc{\JointCond} copilot.}
  \label{tab:real_peg_results}
  \vspace{-10pt}
\end{table}

\vspace{-5pt}
\section{Conclusion}\label{sec:conclusion}
\vspace{-5pt}
We present \alg, a model-free shared autonomy framework that leverages an ODE-based, distilling diffusion model with partial diffusion to realize one-step denoising collaborative control. Without any explicit goal prediction, \alg relies solely on goal-excluded expert demonstrations to “flash back” each user action to its nearest expert distribution—ensuring smooth trajectory corrections while faithfully preserving the user’s intent. In extensive experiments, \alg outperforms prior methods on classic 2D hard-control benchmarks (e.g., Lunar Lander) and delivers powerful assistance in high-dimensional, high-precision tasks such as charger plug insertion—settings where existing approaches falter due to limited data, hyperparameter sensitivity, or slow inference. Moreover, our study of the \JointCond variant uncovers how implicit short-term intention modeling underlies robust fidelity maintenance. These results underscore the promise of consistency-model–based shared autonomy for real-time, goal-agnostic assistance in complex control domains.

\clearpage

\section{Limitations}
While \alg demonstrates strong performance and real-time responsiveness, several areas invite deeper exploration. First, our next-state prediction variant \JointCond can degrade when the surrogate noise level ($\epsilon$) becomes large($\geq 0.5$) under such conditions the surrogate’s outputs may stray from rational human behavior, yielding predictions that add little value. Addressing this will likely require either more robust surrogate models that maintain fidelity across noise regimes or adaptive noise calibration strategies.

Equally important is understanding how the choice of demonstration data shapes assistive performance. Training on “clean,” low-variance expert trajectories may yield highly precise corrections but risks rejecting safe but unconventional user motions; conversely, embracing a broader, higher-variance dataset could improve flexibility yet makes it harder to distinguish core expert maneuvers from benign exploratory actions. Characterizing this trade-off—and developing methods to automatically identify and prioritize “causal” expert patterns—stands as a fertile avenue for future work.

Finally, these two threads converge on the challenge of dynamically tuning the assistance strength ($\alpha$) at inference time. An ideal system would gauge the professionalism of each user action—perhaps via confidence metrics or learned intent classifiers—and adjust $\alpha$ accordingly, providing stronger corrections when needed and stepping back when the user demonstrates competence. We view the development of such context-aware, real-time modulation mechanisms as a highly promising direction to enhance both performance and user trust in consistency-model–based shared autonomy.

\bibliography{references}  

\clearpage
\section{Appendix}

\subsection{Preliminary: Probability Flow ODEs}\label{sec:preliminary-pf-odes}
Diffusion models can be interpreted as discretized numerical solvers for the reverse‑time differential equations that govern a forward diffusion process. A diffusion model is characterized by two processes. A \emph{forward diffusion process} is a first-order Markov chain that iteratively adds noise to a (clean) sample drawn from a data distribution $\bm{y} \sim p_{\mathrm{data}}(\bm{y})$. A \emph{reverse diffusion process} serves to generate samples from the data distribution by reversing this process to denoise a noisy input $\bm{x}$. 
In the general approach, a neural network $D(\cdot)$ is trained to approximate the instantaneous score function $\nabla_x \log p_t(x)$ as a parameterization of the reverse process. By instantiating the generative process using the probability‐flow (PF) ODE, the general form that describes the evolution of a sample (forward or backward in time) can be defined as~\cite{CMmodel,EDM}:
\begin{equation}
    \mathrm{d} \bm{x}=-\dot{\sigma}(t) \sigma(t) \nabla_{\bm{x}} \log p_t(\bm{x} ; \sigma(t)) \mathrm{d} t,
\end{equation}
where $\sigma(t)$ is the desired noise level at time $t$ and in practice setting as $t$ of the EDM model~\cite{EDM} this paper applies.
In the forward process, we represent $p_t(\bm{x} ; \sigma)=p_{\mathrm{data}} \cdot \mathcal{N}\left(\mathbf{0}, \sigma(t)^2 \mathbf{I}\right)$. The denoiser function $D(\bm{x} ; \sigma)$ is designed to minimize the $L^2$ error for samples $\bm{y}$ drawn from $p_{\mathrm{data}}(\bm{y})$
\begin{equation}\label{eq: ODEloss}
    \mathbb{E}_{\bm{y} \sim p_{\text {data }}} \mathbb{E}_{\bm{n} \sim \mathcal{N}\left(\mathbf{0}, \sigma^2 \mathbf{I}\right)} \lVert D(\bm{y}+\bm{n} ; \sigma)-\bm{y} \rVert_2^2,
\end{equation}
where $\bm{n}$ is the noise. The relationship between $D(\bm{x} ; \sigma)$ and the score function is given by
\begin{equation}
    \nabla_{\bm{x}} \log p(\bm{x} ; \sigma)=(D(\bm{x} ; \sigma)-\bm{x}) / \sigma^2.
\end{equation} 
As stated in Section~\ref{sec:relatedwork}, the ODE training objective consider the nearest target distribution in the multi-modal scenario. Expanding Eqn.~\ref{eq: ODEloss} (see Appendix~\ref{Appendix: full eq proof}), the loss becomes

\begin{equation}
    \mathcal{L}(D ;\bm{x}, \sigma) = \int_{\mathbb{R}^d} \frac{1}{Y} \sum_{i=1}^Y \mathcal{N}\left(\bm{x} ; \bm{y}_i, \sigma^2 \mathbf{I}\right)\lVert D(\bm{x} ; \sigma)-\bm{y}_i\rVert_2^2 \mathrm{~d} \bm{x}
\end{equation}

Minimizing $\mathcal{L}(D ;\bm{x}, \sigma)$ is equivalent to solving a convex optimization problem, and the closed-form solution follows as:%
\begin{equation}
    D(\bm{x} ; \sigma)=\frac{\sum_i \mathcal{N}\left(\bm{x} ; \bm{y}_i, \sigma^2 \mathbf{I}\right) \bm{y}_i}{\sum_i \mathcal{N}\left(\bm{x} ; \bm{y}_i, \sigma^2 \mathbf{I}\right)},
\end{equation}

The above involves a softmax operation over all target distributions $\bm{y}_i$ that is feasible to noise sample $\bm{x}$, therefor guarantee to choose the nearest neighbor. Fig.~\ref{fig:2D example} shows that SDE based DDPM model has no guarantee on finding nearest expert, while ODE based model could do this in one step after distilling.

\begin{figure}[!h]
    \centering
    \includegraphics[width=0.65\linewidth]{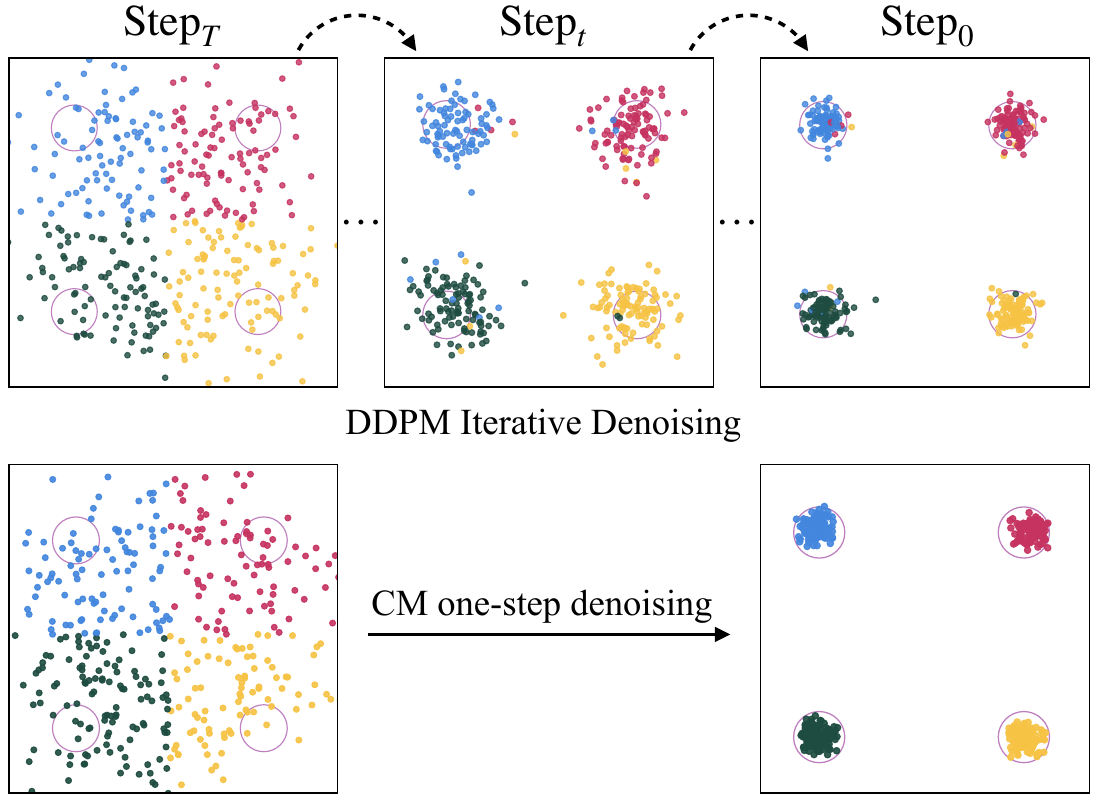}
    \caption{Consistency model and DDPM on a 2D example}
    \label{fig:2D example}
\end{figure}

\subsection{Full proof of nearest expert}\label{Appendix: full eq proof}
Let us assume that our training set consists of a finite number of samples $\left\{\boldsymbol{y}_1, \ldots, \boldsymbol{y}_Y\right\}$. This implies $p_{\text {data }}(\boldsymbol{x})$ is represented by a mixture of Dirac delta distributions:

$$
p_{\text {data }}(\boldsymbol{x})=\frac{1}{Y} \sum_{i=1}^Y \delta\left(\boldsymbol{x}-\boldsymbol{y}_i\right)
$$
which allows us to also express $p(x; \sigma)$ in closed form,
$$
\begin{aligned}
p(\boldsymbol{x} ; \sigma) & =p_{\mathrm{data}} * \mathcal{N}\left(\mathbf{0}, \sigma(t)^2 \mathbf{I}\right) \\
& =\int_{\mathbb{R}^d} p_{\mathrm{data}}\left(\boldsymbol{x}_0\right) \mathcal{N}\left(\boldsymbol{x} ; \boldsymbol{x}_0, \sigma^2 \mathbf{I}\right) \mathrm{d} \boldsymbol{x}_0 \\
& =\int_{\mathbb{R}^d}\left[\frac{1}{Y} \sum_{i=1}^Y \delta\left(\boldsymbol{x}_0-\boldsymbol{y}_i\right)\right] \mathcal{N}\left(\boldsymbol{x} ; \boldsymbol{x}_0, \sigma^2 \mathbf{I}\right) \mathrm{d} \boldsymbol{x}_0 \\
& =\frac{1}{Y} \sum_{i=1}^Y \int_{\mathbb{R}^d} \mathcal{N}\left(\boldsymbol{x} ; \boldsymbol{x}_0, \sigma^2 \mathbf{I}\right) \delta\left(\boldsymbol{x}_0-\boldsymbol{y}_i\right) \mathrm{d} \boldsymbol{x}_0 \\
& =\frac{1}{Y} \sum_{i=1}^Y \mathcal{N}\left(\boldsymbol{x} ; \boldsymbol{y}_i, \sigma^2 \mathbf{I}\right) .
\end{aligned}
$$

Let us now consider the denoising score matching loss of Eq.~\ref{eq: ODEloss}. By expanding the expectations, we can rewrite the formula as an integral over the noisy samples $\boldsymbol{x}$:

$$
\begin{aligned}
\mathcal{L}(D ; \sigma) & =\mathbb{E}_{\boldsymbol{y} \sim p_{\text {data }}} \mathbb{E}_{\boldsymbol{n} \sim \mathcal{N}\left(\mathbf{0}, \sigma^2 \mathbf{I}\right)}\|D(\boldsymbol{y}+\boldsymbol{n} ; \sigma)-\boldsymbol{y}\|_2^2 \\
& =\mathbb{E}_{\boldsymbol{y} \sim p_{\text {data }}} \mathbb{E}_{\boldsymbol{x} \sim \mathcal{N}\left(\boldsymbol{y}, \sigma^2 \mathbf{I}\right)}\|D(\boldsymbol{x} ; \sigma)-\boldsymbol{y}\|_2^2 \\
& =\mathbb{E}_{\boldsymbol{y} \sim p_{\text {data }}} \int_{\mathbb{R}^d} \mathcal{N}\left(\boldsymbol{x} ; \boldsymbol{y}, \sigma^2 \mathbf{I}\right)\|D(\boldsymbol{x} ; \sigma)-\boldsymbol{y}\|_2^2 \mathrm{~d} \boldsymbol{x} \\
& =\frac{1}{Y} \sum_{i=1}^Y \int_{\mathbb{R}^d} \mathcal{N}\left(\boldsymbol{x} ; \boldsymbol{y}_i, \sigma^2 \mathbf{I}\right)\left\|D(\boldsymbol{x} ; \sigma)-\boldsymbol{y}_i\right\|_2^2 \mathrm{~d} \boldsymbol{x} \\
& =\int_{\mathbb{R}^d} \underbrace{\frac{1}{Y} \sum_{i=1}^Y \mathcal{N}\left(\boldsymbol{x} ; \boldsymbol{y}_i, \sigma^2 \mathbf{I}\right)\left\|D(\boldsymbol{x} ; \sigma)-\boldsymbol{y}_i\right\|_2^2}_{=: \mathcal{L}(D ; \boldsymbol{x}, \sigma)} \mathrm{d} \boldsymbol{x} .
\end{aligned}
$$

The proof is adopted from EDM appendix~\cite{EDM}.

\subsection{Surrogate hyperparameter choose and performance}\label{Appendix: surrogate performance}

In Fig.~\ref{fig: Surrogate Lunar Lander}, we can see for noised surrogate, when $\epsilon=0.0$, the surrogate is actually the expert, and when $\epsilon=0.45$ (Fig.~\ref{fig: Noised Surrogate Lunar Lander}) the success rate drop down significantly to 20\%, so we choose $\epsilon^{\mathrm{noised}}=0.45$ for lunar lander environment. Similarly, we can so the same choosing for peg insertion and charger insertion, table \ref{tab:surrogate-flaws} shows the hyperparameter $\epsilon$ we use.

\begin{figure}[!h]
  \centering
  \begin{subfigure}[b]{0.25\textwidth}
    \centering
    \includegraphics[width=\linewidth]{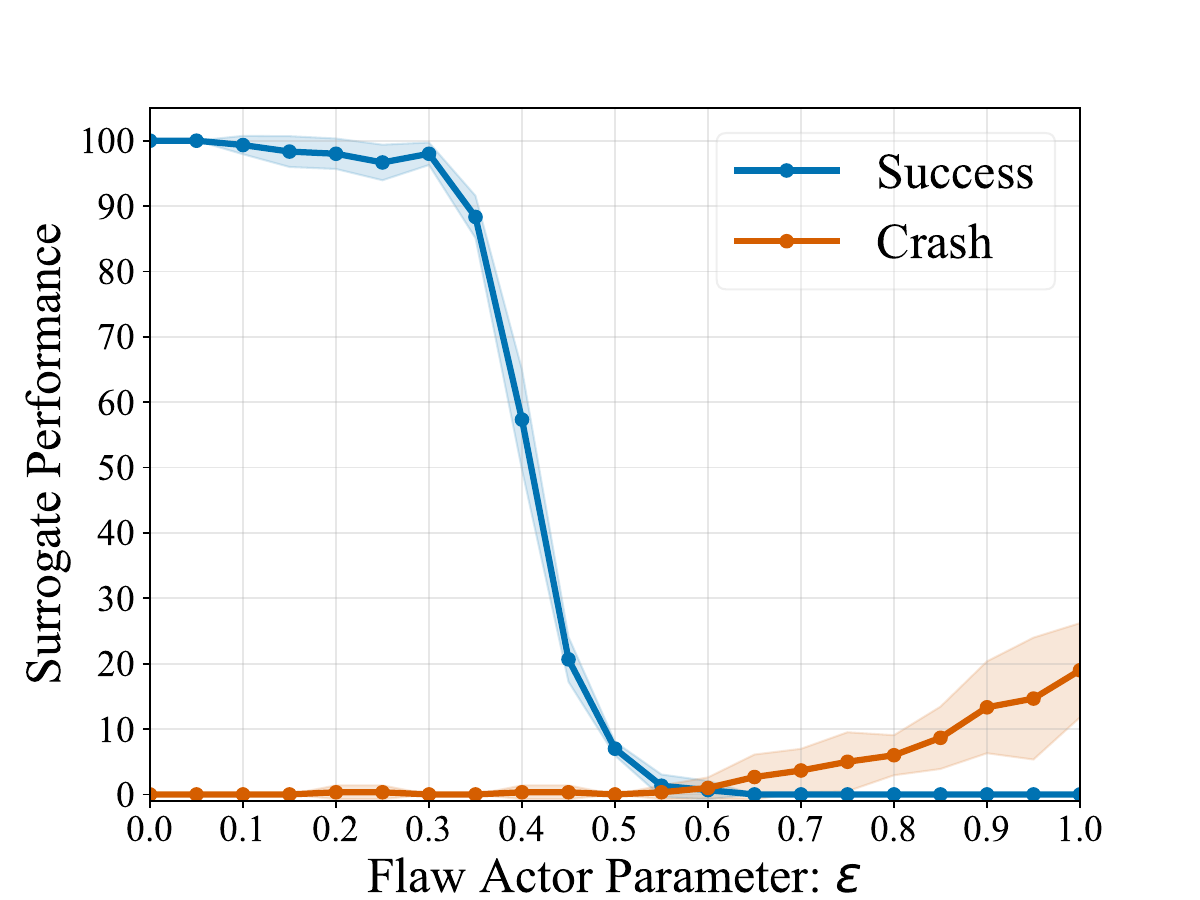}
    \caption{Noised Surrogate}
    \label{fig: Noised Surrogate Lunar Lander}
  \end{subfigure}%
  \hfill
  \begin{subfigure}[b]{0.25\textwidth}
    \centering
    \includegraphics[width=\linewidth]{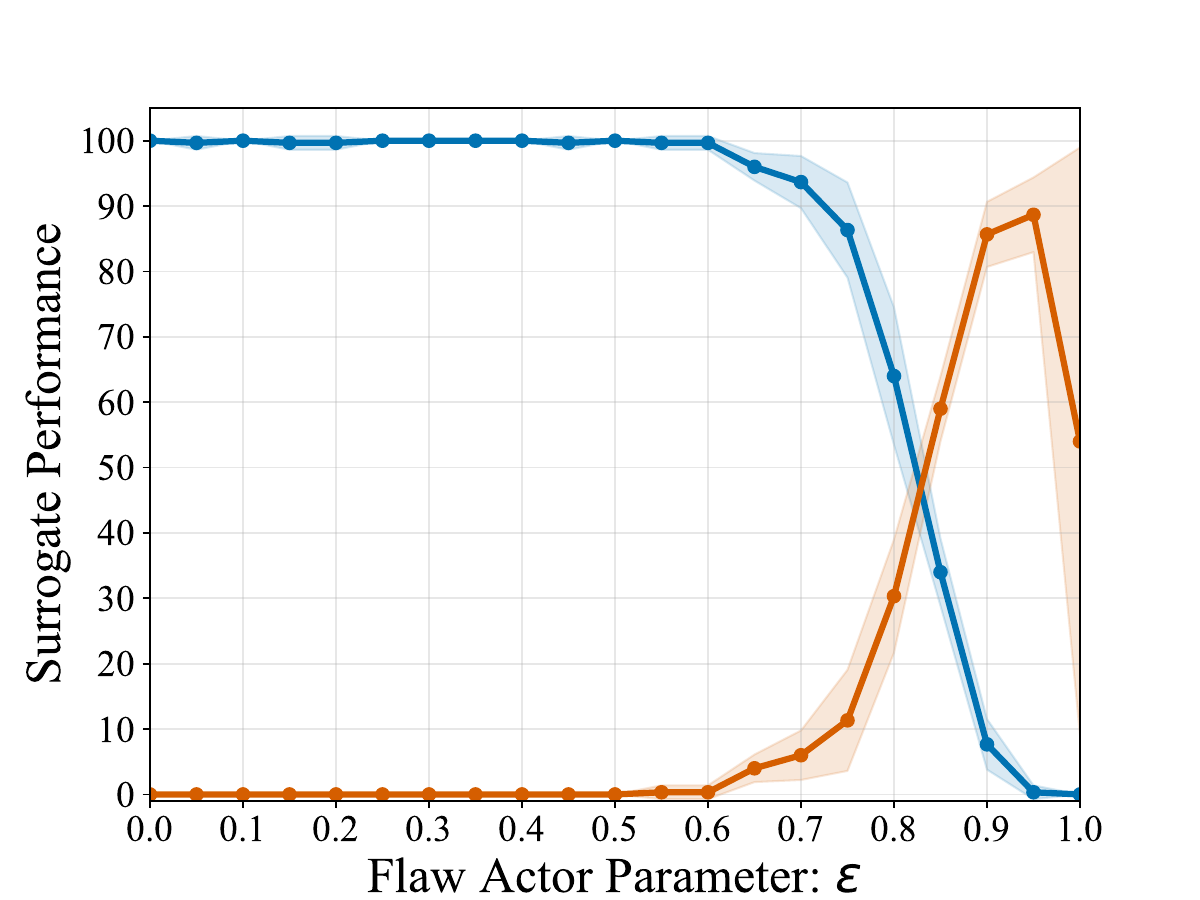}
    \caption{Laggy Surrogate}
    \label{fig:Laggy Surrogate Lunar Lander}
  \end{subfigure}%
  \hfill
  \begin{subfigure}[b]{0.25\textwidth}
    \centering
    \includegraphics[width=\linewidth]{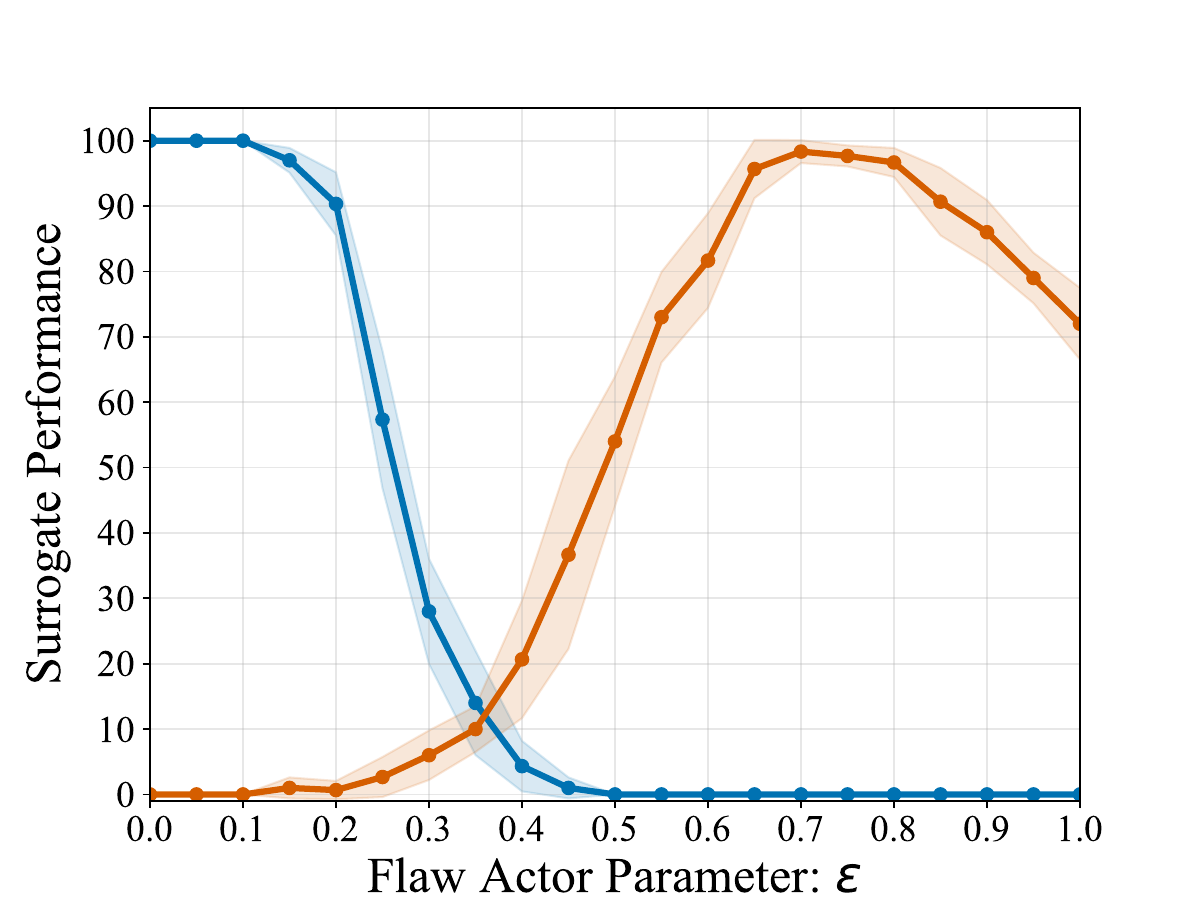}
    \caption{Noisy Surrogate}
    \label{fig: Noisy Surrogate Lunar Lander}
  \end{subfigure}%
  \hfill
  \begin{subfigure}[b]{0.25\textwidth}
    \centering
    \includegraphics[width=\linewidth]{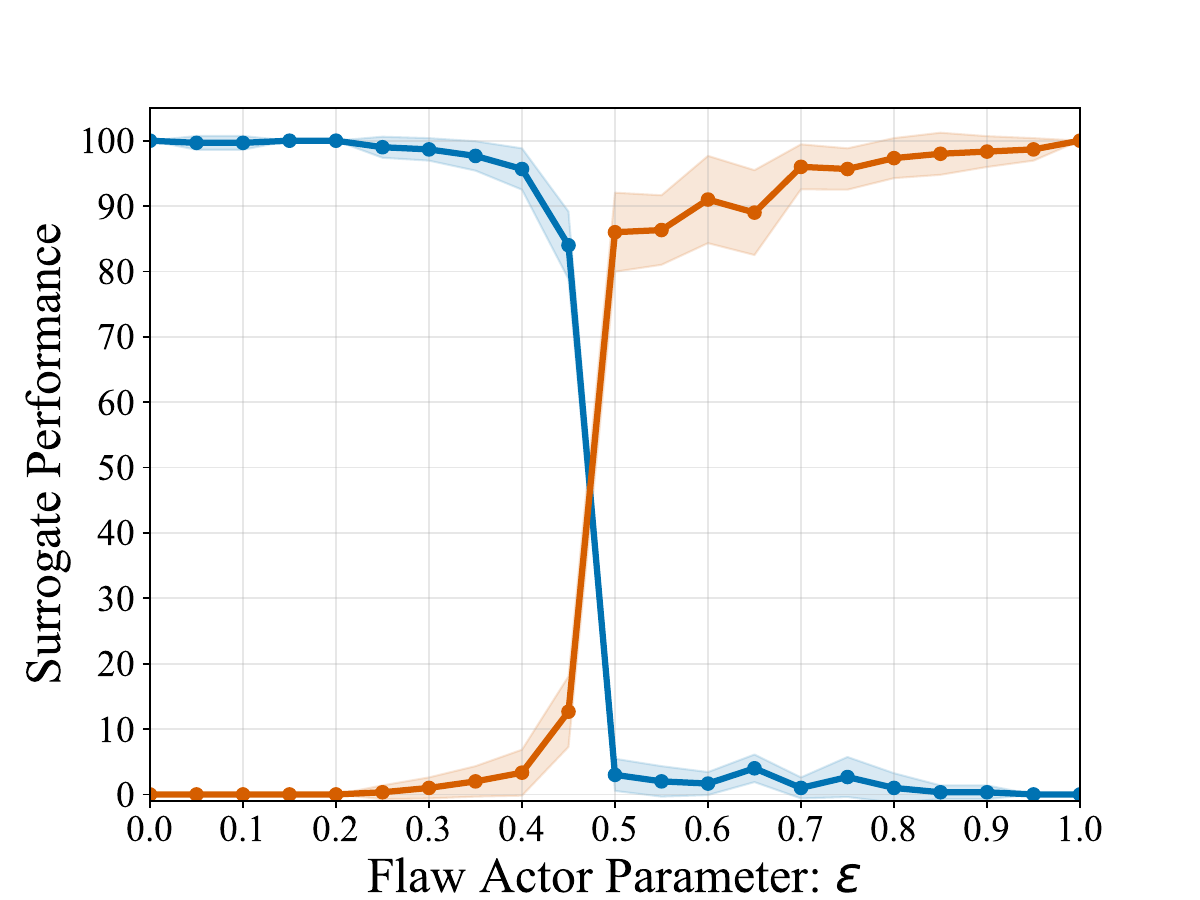}
    \caption{Slow Surrogate}
    \label{fig: Slow Surrogate Lunar Lander}
  \end{subfigure}
  \caption{Surrogate performance over different flaw parameter $\epsilon$ for lunar lander}
  \label{fig: Surrogate Lunar Lander}
\end{figure}

\begin{table}[htbp]
  \centering
  \caption{Surrogate Flaw Parameters}
  \label{tab:surrogate-flaws}
  \begin{tabular}{lcccc}
    \toprule
    Environment & $\epsilon^{\mathrm{laggy}}$ & 
                  $\epsilon^{\mathrm{noisy}}$ & 
                  $\epsilon^{\mathrm{noised}}$ & 
                  $\epsilon^{\mathrm{slow}}$ \\
    \midrule
    Lunar Lander            & 0.85 & 0.30 & 0.45 & 0.47 \\
    Peg Insertion           & 0.80 & 0.15 & 0.20 & 0.65 \\
    Charge Plug Insertion   & 0.60 & 0.15 & 0.10 & 0.40 \\
    \bottomrule
  \end{tabular}
\end{table}

\subsection{EDM Details}\label{Appendix: EDM hyperparameter}

An EDM model is fully specified once the \emph{noise schedule}
\(\{\sigma_t\}_{t=0}^{T-1}\) is fixed.
The schedule assigns a noise standard deviation \(\sigma_t\) to each discrete
diffusion step \(t\).
Throughout this paper we adopt the default hyperparameters of \citet{EDM}:
\[
\sigma_{\min}=0.002,\qquad
\sigma_{\max}=80.0,\qquad
\sigma_{\text{data}}=0.5,\qquad
\rho=7.
\]

\paragraph{Karras \(\rho\)-schedule.}
For \(T\) noise levels and index \(t\in\{0,\dots,T-1\}\),  
the schedule is
\[
\sigma^t \;=\;
\Bigl(
   \sigma_{\max}^{1/\rho}
   +\frac{t}{T-1}\,
    \bigl(\sigma_{\min}^{1/\rho}-\sigma_{\max}^{1/\rho}\bigr)
\Bigr)^{\rho}.
\tag{A.1}
\]

\paragraph{Network pre-conditioning coefficients.}
At any noise level \(\sigma^t\), EDM rescales the network input, skip
connection, and output using
\[
\begin{aligned}
c_{\text{in}}^t   &= \frac{1}{\sqrt{(\sigma^{t})^{2}+\sigma_{\text{data}}^{\,2}}},\\[4pt]
c_{\text{skip}}^t &= \frac{\sigma_{\text{data}}^{\,2}}
                        {(\sigma^{t})^{2}+\sigma_{\text{data}}^{\,2}},\\[4pt]
c_{\text{out}}^t  &= \frac{\sigma^t\,\sigma_{\text{data}}}
                        {\sqrt{(\sigma^{t})^{2}+\sigma_{\text{data}}^{\,2}}}.
\end{aligned}
\tag{A.2}
\]

These coefficients stabilise both training and sampling across the wide
dynamic range between \(\sigma_{\max}\) and \(\sigma_{\min}\).

\paragraph{Sampling a diffusion time step.}
Whenever a single training batch requires a noise level, we draw the
index \(t\) \emph{uniformly} from the schedule:
\[
t\;\sim\;\mathcal{U}\bigl\{0,\dots,T-1\bigr\},
\qquad
\sigma \;\leftarrow\; \sigma_t .
\]

\subsection{Experimental Details}\label{sec:appendix-experimental-details}
\emph{(a) Lunar Lander}: Lunar Lander (Fig.~\ref{fig: Lunar Lander}) is a 2D continuous control environment adapted from OpenAI Gym, where the goal is to land a spaceship on a designated landing pad. The action space controls the thrust applied to the left and right engines; to turn left, for example, the agent must apply slightly more thrust to the right engine than to the left. This indirect control strategy is counterintuitive compared to natural human intuition (e.g., pushing left to move left), making the task particularly challenging. The state space includes the spaceship's position, orientation, linear and angular velocities, ground contact indicators for each leg, and the landing pad location (the latter provided only to the pilot). An episode terminates when the spaceship successfully lands and becomes idle, crashes, flies out of bounds, or reaches a timeout of 1000 steps.

\emph{(b) Peg Insertion}: Peg Insertion (Fig.~\ref{fig: Peg Insertion}) is a continuous control environment from ManiSkill, involving a robotic arm tasked with inserting the orange end of an orange-white peg into a hole in a box. The peg has a fixed half-length of 10 cm and radius of 2 cm, while the box hole radius provides a 1 cm clearance. At episode start, both peg and box positions and orientations are randomized on a flat table surface. Actions involve continuous end-effector movements. The state includes robot joint positions and velocities, end-effector pose, peg pose and dimensions, and the hole's pose and radius. Success occurs when the peg’s white end is within 1.5 cm of the hole center. Episodes end upon success, boundary violation, or reaching a 200-step limit.
\begin{figure*}[!t]
  \centering
  \begin{subfigure}[b]{0.32\textwidth}
    \includegraphics[width=\textwidth]{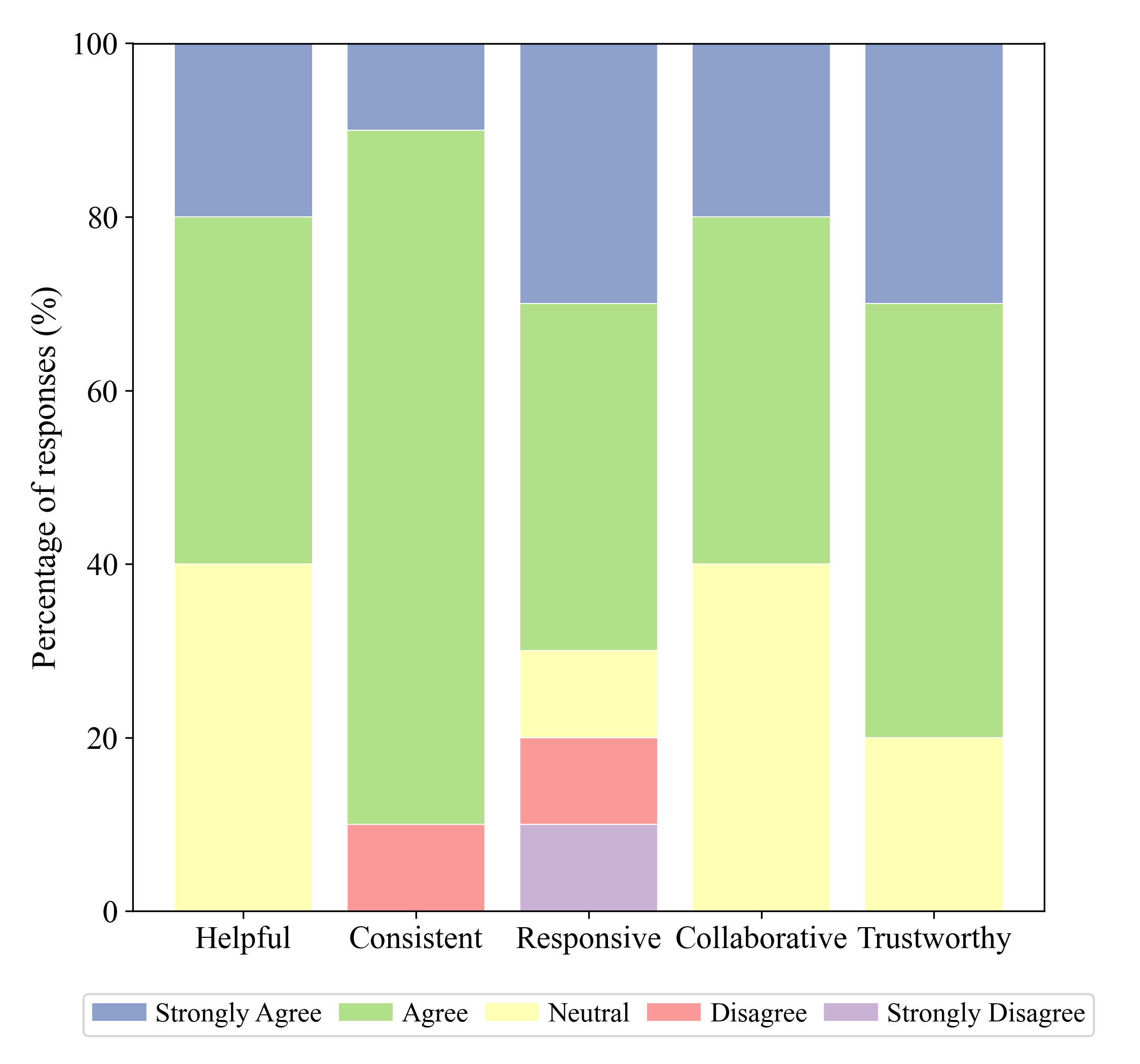}
    \caption{Responses with Assistive Policy}
    \label{fig:assistive-response}
  \end{subfigure}
  \hfill
  \begin{subfigure}[b]{0.32\textwidth}
    \includegraphics[width=\textwidth]{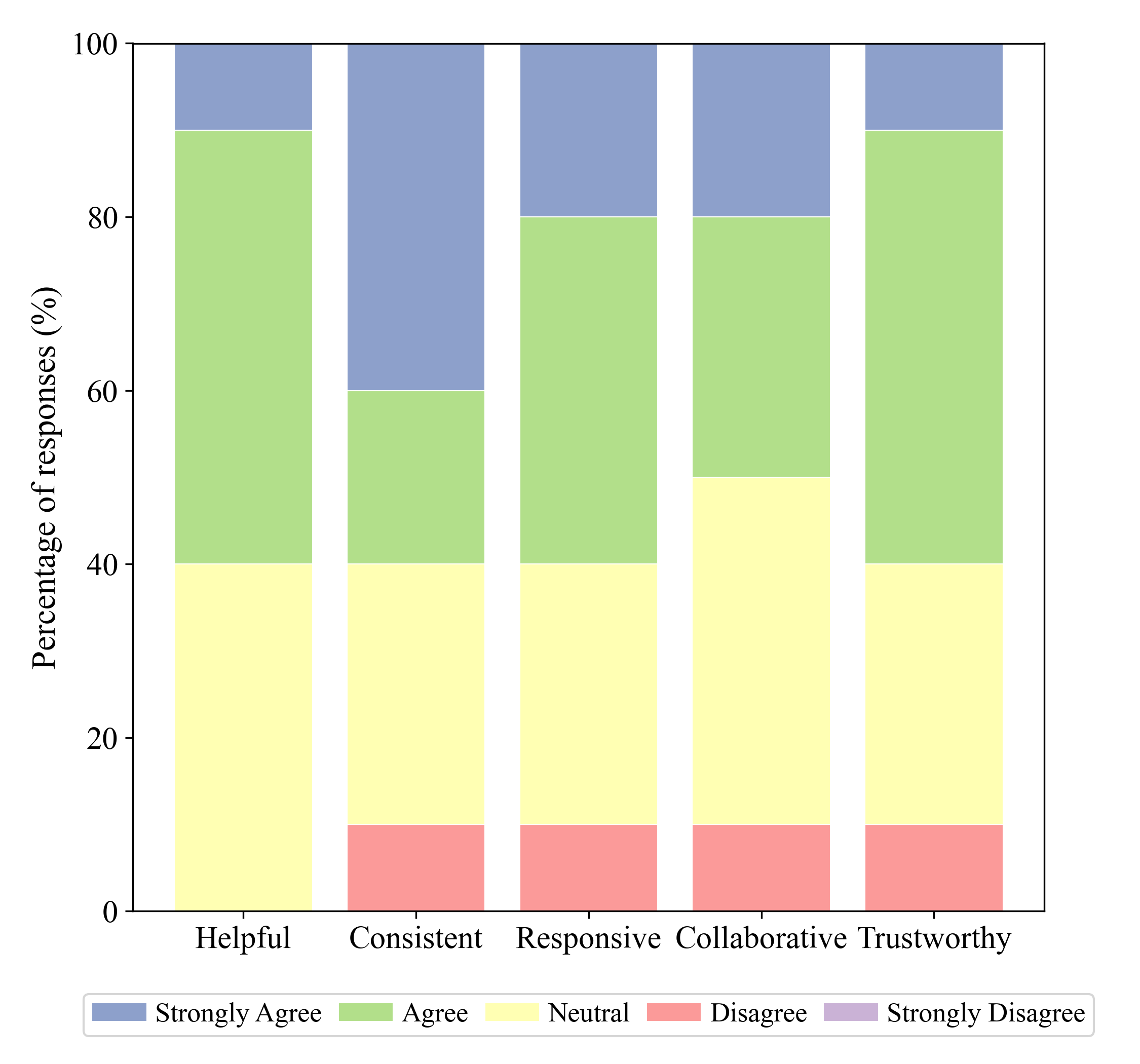}
    \caption{Responses with Teleoperation}
    \label{fig:teleop-response}
  \end{subfigure}
  \hfill
  \begin{subfigure}[b]{0.32\textwidth}
    \includegraphics[width=\textwidth]{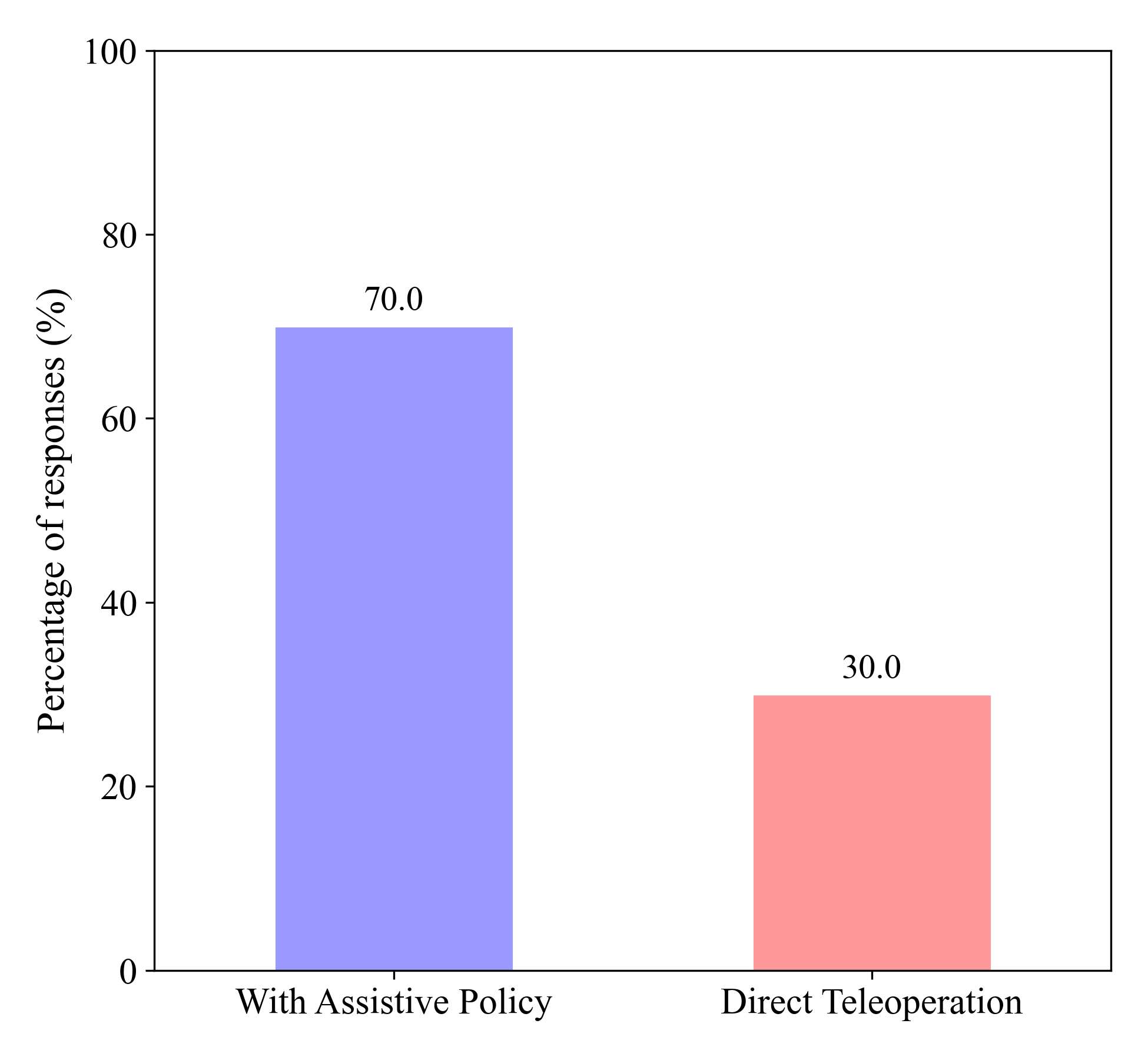}
    \caption{Participant Overall Preference}
    \label{fig:overall-preference}
  \end{subfigure}
  \caption{Human participant qualitative survey result in the Real Peg Insertion task}
  \label{fig: human-survey-real-peg}
\end{figure*}

\emph{(c) Charger Plug Insertion}: Charger plug (Fig.~\ref{fig: Charger Plug Insertion}) insertion is an advanced version of peg insertion where a robotic arm must pick up a charger and insert it into a receptacle. The charger has a base of fixed size and a dual-peg design, with a clearance of \emph{0.5 mm} for insertion. At episode start, the charger and receptacle are randomized in XY position and orientation on the table. Actions control the robot's end-effector. The state includes robot joint positions and velocities, end-effector pose, charger pose, receptacle pose, and goal pose. Success is achieved when the charger is fully inserted within a \emph{5mm} positional tolerance and 0.2 radian angular tolerance. Episodes end upon success, boundary violation, or reaching a 300-step limit.

\emph{(d) Real Peg Insertion}: Real Peg Insertion (Fig.~\ref{fig: Real Peg Insertion}) is performed with a UR5 robot arm equipped with a Robotiq 3-Finger gripper. The task involves inserting either a red square peg or a blue cylinder peg into its corresponding hole, with clearance tolerances of 5\,mm and 4.5\,mm, respectively.   
The state~$s_t$ includes the current pose of the end-effector and the wrench readings from a wrist-mounted FT-300 force/torque sensor.
The action~$a_t$ is the target end-effector position.

\begin{table}[!t]
  \centering
  \label{tab:env_dims}
  \begin{tabular}{lcc}
    \toprule
    Environment & State Dimension & Action Dimension \\
    \midrule
    Lunar Lander (Fig.~\ref{fig: Lunar Lander})          & 8  & 2 \\
    Peg Insertion (Fig.~\ref{fig: Peg Insertion})        & 35 & 8 \\
    Charger Plug Insertion (Fig.~\ref{fig: Charger Plug Insertion}) & 32 & 8 \\
    Real Peg Insertion (Fig.~\ref{fig: Real Peg Insertion})       & 9 & 3 \\
    \bottomrule
  \end{tabular}
  \vspace{1.5mm}
  \caption{State and Action Dimensions for Each Environment}
\end{table}

\subsection{Human Participant Qualitative Survey Result} \label{subsec: human-survey-result}

In a qualitative survey, we asked participants to rate how “helpful,” “consistent,” “responsive,” “collaborative,” and “trustworthy” each copilot felt, using a five-point Likert scale. They were also asked which copilot they preferred overall. To avoid bias, we did not reveal which system was our assistive policy; the survey simply labeled them “Policy A” and “Policy B.”

As shown in Fig.~\ref{fig: human-survey-real-peg}, participants rated our assistive copilot higher for being helpful, collaborative, and trustworthy, and the majority preferred it overall.
We also received comments such as “The manipulation and motion are smoother under Policy B (our assistive policy) from the user’s perspective,” “It felt like I got some assistance during insertion with B,” and “Policy A (direct teleoperation) is easy to overshoot, especially when moving down.”

\end{document}